\documentclass{article}

\usepackage[final]{neurips_2025}

\usepackage[utf8]{inputenc} 
\usepackage[T1]{fontenc}    
\usepackage{url}            
\usepackage{booktabs}       
\usepackage{amsfonts}       
\usepackage{nicefrac}       
\usepackage{microtype}      
\usepackage{xcolor}         

\usepackage{microtype}
\usepackage{graphicx}
\usepackage{subfigure}
\usepackage{booktabs} 
\usepackage{multirow}
\usepackage[draft]{minted}

\usepackage{amsmath}
\usepackage{amssymb}
\usepackage{mathtools}
\usepackage{amsthm}
\usepackage{makecell}
\usepackage{wrapfig}
\usepackage{appendix}
\usepackage{titletoc}
\usepackage{changepage}  

\usepackage{longtable}

\definecolor{citecolor}{HTML}{0071BC}
\definecolor{linkcolor}{HTML}{ED1C24}
\definecolor{pearDark}{HTML}{2980B9}

\usepackage{pifont}
\usepackage[misc]{ifsym}

\usepackage[pagebackref=false,breaklinks=true,colorlinks,citecolor=citecolor,linkcolor=linkcolor,bookmarks=false]{hyperref}

\title{GEM: Empowering MLLM for Grounded ECG Understanding with Time Series and Images}

\author{
  Xiang Lan$^{1}$, Feng Wu$^{1}$, Kai He$^{1}$, Qinghao Zhao$^{2}$, Shenda Hong$^{3}$\textsuperscript{\Letter}, Mengling Feng$^{1}$\textsuperscript{\Letter}\\
  $^{1}$National University of Singapore
  $^{2}$Peking University People’s Hospital
  $^{3}$Peking University \\
  \textsuperscript{\Letter}Correspondence to: \texttt{hongshenda@pku.edu.cn}, \texttt{mornin@nus.edu.sg}
}

\begin{document}

\maketitle

\begin{abstract}

While recent multimodal large language models (MLLMs) have advanced automated ECG interpretation, they still face two key limitations: (1) insufficient multimodal synergy between ECG time series and ECG images, and (2) limited explainability in linking diagnoses to granular waveform evidence. We introduce GEM, the first MLLM unifying ECG time series, 12-lead ECG images and text for grounded and clinician-aligned ECG interpretation. GEM enables feature-grounded analysis, evidence-driven reasoning, and a clinician-like diagnostic process through three core innovations: a dual-encoder framework extracting complementary time series and image features, cross-modal alignment for effective multimodal understanding, and knowledge-guided instruction data generation for generating high-granularity grounding data (ECG-Grounding) linking diagnoses to measurable parameters ($e.g.$, QRS/PR Intervals). Additionally, we propose the Grounded ECG Understanding task, a clinically motivated benchmark designed to comprehensively assess the MLLM's capability in grounded ECG understanding. Experimental results on both existing and our proposed benchmarks show GEM significantly improves predictive performance (CSN $7.4\% \uparrow$), explainability ($22.7\% \uparrow$), and grounding ($25.3\% \uparrow$), making it a promising approach for real-world clinical applications.
Codes, model, and data are available at \href{https://github.com/lanxiang1017/GEM}{https://github.com/lanxiang1017/GEM}.

\end{abstract}

\begin{figure*}[!h]
    \centering
    \includegraphics[width=\textwidth]{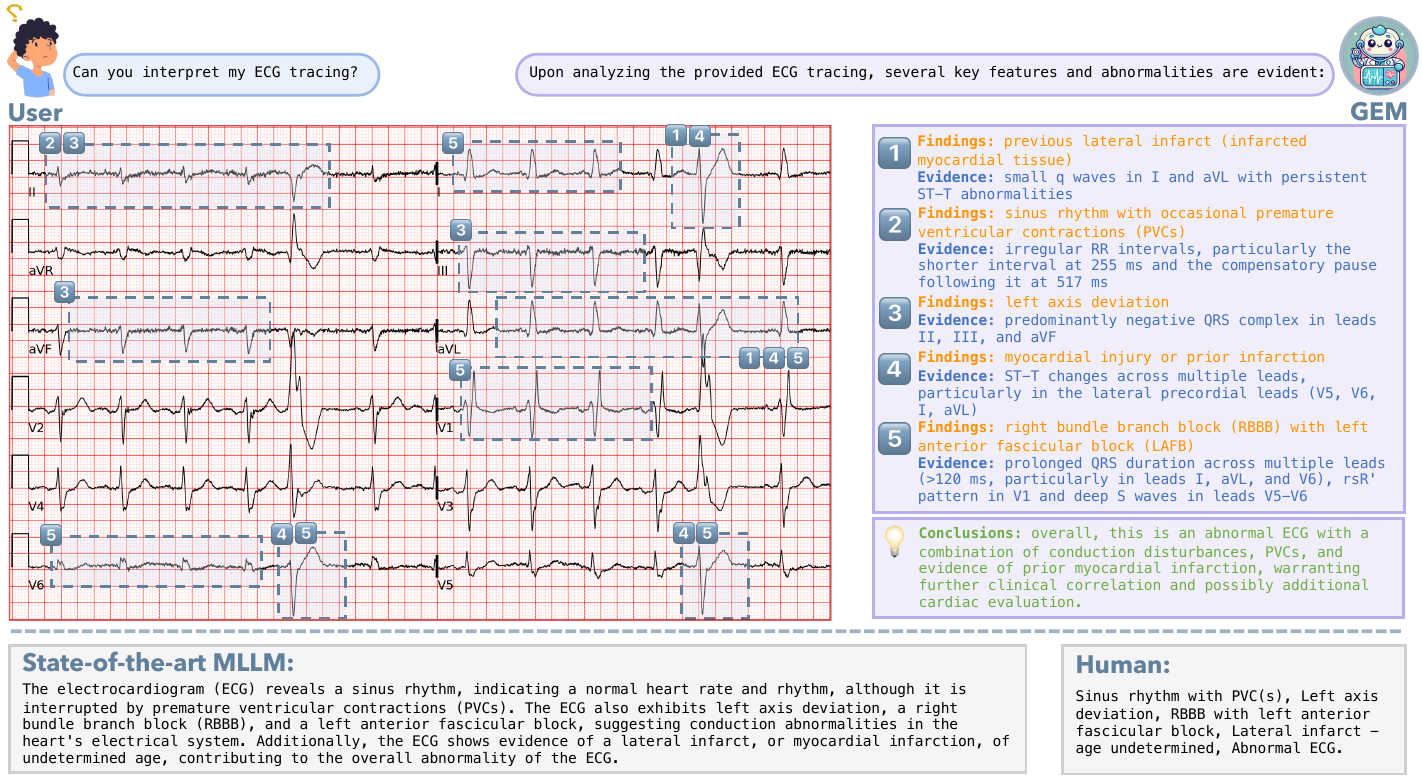}
    \caption{
    GEM offers superior granularity in ECG interpretation compared to state-of-the-art models and human-written reports.
    }
    \label{fig:introfig.}
\end{figure*}
\label{abstract}

\section{Introduction}
\label{introduction}
Electrocardiography (ECG), a cornerstone of cardiac diagnostics, captures the heart's electrical activity through body-surface electrodes, enabling non-invasive assessment of cardiac physiology and pathology \citep{berkaya2018survey, hannun2019cardiologist}. Clinical ECG interpretation synergizes computational and clinical expertise: automated algorithms process raw signals to detect fiducial points ($e.g.$, R-wave peaks) and generate diagnostic hypotheses, while clinicians validate these findings through 12-lead waveform analysis \citep{SMULYAN2019153}. By contextualizing algorithmic outputs with patient-specific factors, clinicians resolve ambiguities, detect subtle anomalies, and formulate diagnoses. This synergy between computational precision and expert judgment ensures reliable and holistic diagnoses in clinical practice.

Deep learning models have achieved promising results in cardiac anomalies detection \citep{hong2020opportunities, zhu2021identification, kiyasseh2020clocs, lan2022intra, yang2023biot, zhao2024deep, pmlr-v227-li24a, lan2024towards, mckeen2024ecg} 
yet lack language capability and model explainability. While recent MLLMs like PULSE \citep{liu2024teach} have advanced language-based ECG interpretation via large-scale instruction tuning ($e.g.$, ECG-Instruct’s 1M+ samples), they primarily focus on static image inputs and predefined diagnostic tasks, leaving two critical gaps: \textbf{Limited Modality Exploration}. Current models typically process only one non-text modality, overlooking the benefits of a synergistic approach that combining time series and image-based modeling. For example, time series models capture dynamic changes but may overlook spatial patterns, while image-based models detect global structures but may miss subtle temporal details. This limits their ability to replicate clinicians' holistic reasoning, which integrates both machine-measured temporal signals and waveform patterns from 12-lead plots. \textbf{Insufficient Explainability and Grounding}. Most existing models designed for ECG provides limited explainability, failing to explicitly connect diagnoses to granular waveform evidence and provide insight into their diagnostic reasoning. A trustworthy ECG model should not only predict cardiac conditions but also explicitly highlight which ECG features led to those conclusions. Such grounded explanations enhance transparency and make model outputs more reliable and actionable for clinical decision-making.

In this work, we introduce GEM, a multimodal large language model designed for grounded ECG understanding by integrating time series, image, and text data. As a conversational cardiology AI assistant, GEM differs from other MLLMs through three key features. First, it provides \textbf{\textit{feature-grounded analysis}}, ensuring that its findings are explicitly tied to detailed ECG features. Second, it offers \textbf{\textit{evidence-driven diagnosis}}, where its conclusions are supported by clear and logical reasoning directly linked to ECG findings. Lastly, GEM simulates a \textbf{\textit{realistic interpretation process}}, mimicking how a clinician analyzes ECGs and arrive at a diagnosis. 

Achieving these capabilities entails challenges in two key dimensions. On the modeling side, it is crucial to effectively integrate information from different modalities to support accurate ECG understanding. On the data side, there is currently no available instruction data designed for training LLMs on high-granularity ECG interpretation.

We tackle these challenges through three novel approaches.
\textit{Multimodal Encoding} allows GEM to extract and integrate complementary features from raw ECG time series and their transformed images. It employs a dual-encoder architecture, with each encoder specialized in its respective modality using established models from the time series and vision domains. This design leverages the unique strengths of both modalities.
\textit{Cross-modal Alignment Learning} facilitates the interpretation of multimodal ECG data by the LLM. Time series representations are first projected to image representations dimensionality, followed by a shared projector that transforms both into language-like embeddings that are comprehensible to LLM. These aligned embeddings are then fused with textual instruction embeddings, enabling effective multimodal understanding through next-token prediction training.
\textit{Knowledge-guided Instruction Data Generation} supports the construction of high-granularity instruction data annotated with heartbeat-level physiological features, without the need for manual annotation. This methodology integrates a grounding feature extractor, which derives precise physiological features from ECG time series, and a cardiology-specific diagnosis guider, which processes these features into structured prompts to more effectively leverage GPT-4o’s latent medical knowledge for generating clinically detailed and feature-grounded ECG instruction data.
Ultimately, GEM delivers significantly more detailed and informative interpretations than both human-written reports and leading MLLMs, as shown in Figure \ref{fig:introfig.}.

The main contributions of this work are three-folds:

\textit{1.First Unified Multimodal ECG Model}. 
We present GEM, the first multimodal framework to synergistically integrate raw ECG time seriesa, 12-lead ECG plots, and textual instructions, leveraging their complementary strengths to advance grounded ECG understanding.

\textit{2.First High-granularity ECG Grounding Dataset}. 
We propose a novel knowledge-guided instruction data generation method, resulting in ECG-Grounding, a dataset comprising 30,000 instruction pairs annotated with heartbeat-level physiological features. This is the first high-granularity ECG grounding dataset, enabling evidence-based diagnosis and improving the trustworthiness of medical AI.

\textit{3.Clinically Oriented Diagnostic System}. 
We introduce the Grounded ECG Understanding task, a clinically motivated benchmark designed to comprehensively assess a model’s ECG interpretation capability. Experimental results demonstrate that GEM not only excels in predictive performance but also significantly enhances explainability and grounding, making it more applicable for real-world clinical settings while fostering greater trust among medical professionals. 

\section{Related Work}
\label{related}
\subsection{Multimodal Large Language Models}
Large Language Models (LLMs), such as GPTs~\citep{achiam2023gpt}, LLaMA~\citep{touvron2023llama}, have made significant advancements in artificial intelligence. Despite their superior performance on numerous natural language processing tasks, LLMs are inherently limited to the text modality, making them "blind" to other modalities such as images, audio and video. To mitigate this constraint, Multimodal Large Language Models have been recently developed to extend the ability of LLMs in comprehending multiple modalities ~\citep{liu2024visual, 10445007, zhang2025r1, huang2025visual}.
By integrating LLMs with various data sources, MLLMs enable the handling of diverse information beyond text.
For example, LLaVA~\citep{liu2024visual} enables LLMs to comprehend visual inputs by adopting a learnable projector to map image features into the word embedding space. Video-LLaMA~\citep{zhang2023video} further enhances LLMs by enabling video perception and understanding.
Qwen-Audio~\citep{chu2023qwen} introduces an audio-language model capable of processing various audio types, including human speech, natural sounds, and music. Medical data, by nature, are inherently multimodal, encompassing diverse formats such as images, physiological time series, and textual reports. These modalities collectively form the foundation for clinical decision-making, driving the advancement of AI in medical applications 
\citep{li2023llavamed,moor2023foundation,radhakrishnan2023cross,liu2023medical,hong2023simplenomo,zhu2024multimodal,ren2024moving,sellergren2025medgemma,yang2025ecg,jin2025uniecg,HE2025102963}. 
Different from these works, our focus is on empowering MLLMs with the grounded ECG understanding capability.

\subsection{Language-based ECG Analysis}
Language-based ECG diagnosis and interpretation is still in its early stages of development. Only a few recent studies have explored LLM-based approaches for ECG analysis. For instance,
\citet{yu2023zero} proposes a zero-shot retrieval-augmented diagnosis technique, embedding domain knowledge from textbooks and research papers into a vector database to improve zero-shot diagnostic accuracy. \citet{cai2023jolt} proposes JoLT, a framework that jointly models ECG time series and text using a Querying Transformer to align their representations. \citet{liu2024teach} introduces PULSE, an LLM-based framework designed to enhance ECG image understanding for diagnosis and report generation. PULSE synthesizes realistic ECG images from raw ECG signals, enabling better utilization of image-based LLaVA models. \citet{zhao2024ecg} develops ECG-CoCa, an ECG encoder trained on ECG-text pairs, alongside ECG-Chat, a modified LLaVA model capable of processing ECG time series. \citet{chan2024leveraging} proposes an analytical framework integrating time series data with LLMs, combining physiological signal analysis with contextual textual information. \citet{wan2024electrocardiogram} designs an instruction-tuning framework for automated ECG report generation, converting ECG-text pairs into chatbot-style instructions and fine-tuning the LLM’s linear layers.

GEM distinguishes itself from existing language-based ECG models in two key aspects. First, in model architecture, unlike existing methods that limit analysis to isolated modalities (either ECG signals or transformed ECG images) with text, GEM introduces a unified architecture that integrates both time series data and 12-lead images. This approach mirrors a clinician’s natural workflow, where dynamic signal trends and spatial waveform patterns are jointly analyzed for a more comprehensive interpretation. Second, in ECG understanding, GEM establishes a new paradigm for evidence-driven diagnosis. While current models often lack grounded understanding, GEM enables heartbeat-level interpretability by directly linking each diagnostic conclusion to quantifiable physiological evidence, enhancing explainability and clinical reliability. By combining the two paradigm-shifting innovations, GEM addresses fundamental limitations in existing models and advances language-based ECG analysis, setting a new standard for conversational AI-assisted cardiac diagnostics.

\section{Method}
\label{method}
\begin{figure*}[t]
    \centering
    \includegraphics[width=\textwidth]{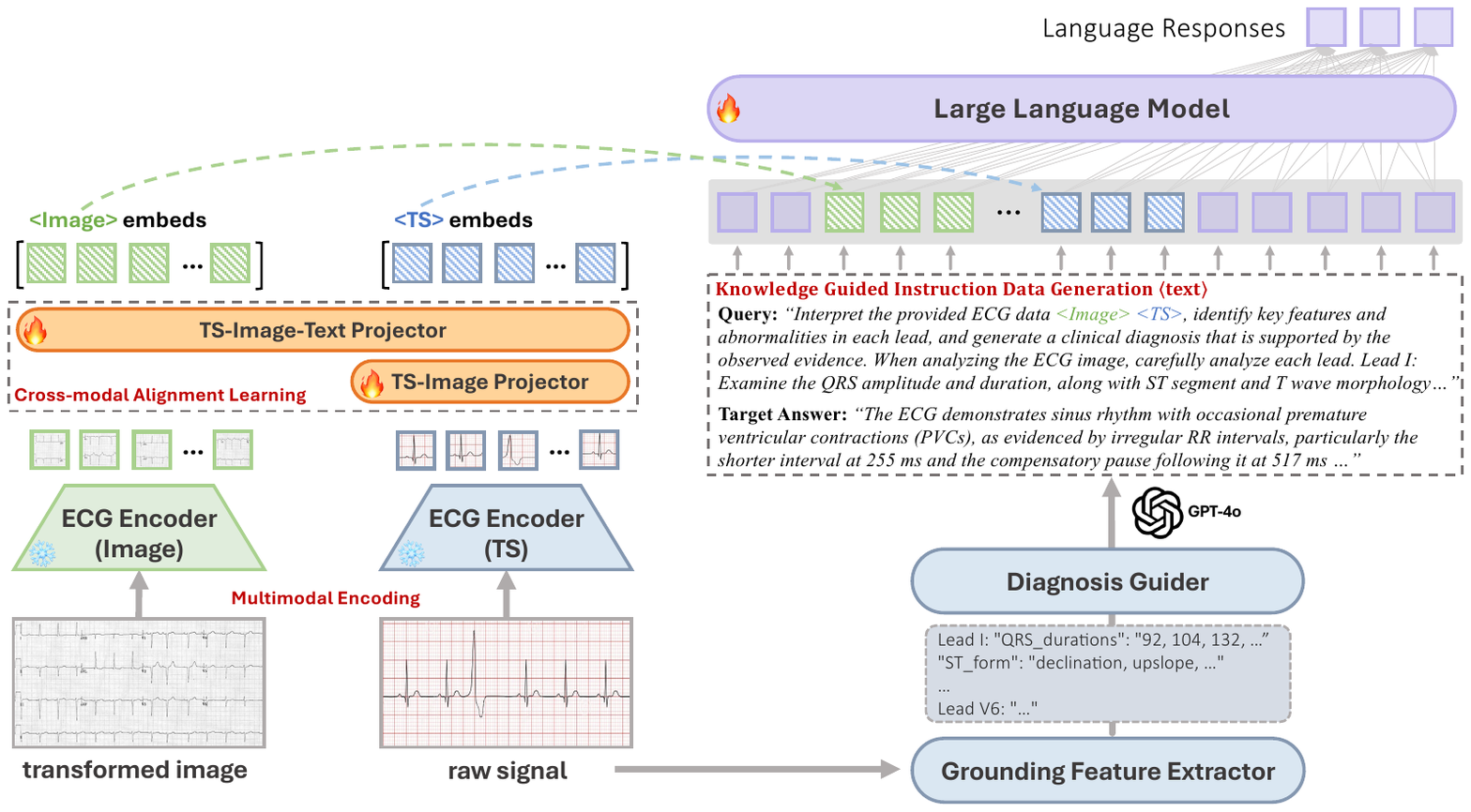}
    \caption{GEM's Architecture.
    \textit{Multimodal Encoding}: Separate encoders process ECG time series and images to generate modality-specific representations, enabling a holistic analysis of ECG data.
    \textit{Cross-modal Alignment Learning}: Time series and image representations are first aligned and then mapped to a textual space using a shared projector, ensuring coherent understanding for the LLM.
    \textit{Knowledge-guided Instruction Data Generation}: Physiological features extracted from all 12 leads are sequenced and structured using a diagnosis guider, which prompts GPT-4o with domain-specific instructions to generate high-granularity instructional data.
    }
    \label{fig:GEM}
\end{figure*}

\subsection{Overview}
GEM's training primarily relies on three key components: the multimodal encoding, the cross-modal alignment learning, and the knowledge-guided instruction data generation. Figure \ref{fig:GEM} provides an overview of the GEM model. We will elaborate on each components in the following sections.

\subsection{Multimodal Encoding}
We represent the ECG time series as $\boldsymbol{x}_{ts} \in \mathbb{R}^{C \times L}$, where $C$ is the number of leads in multi-lead ECG and $L$ is the length of the signal. The transformed ECG image derived from the time series is denoted as $\boldsymbol{x}_{img} \in \mathbb{R}^{H \times W \times 3}$, where $H$ and $W$ are the height and width of the image, respectively. For ECG time series encoder $E_{ts}(\theta_{ts})$, we adopt the pre-trained ECG-CoCa model \citep{zhao2024ecg}, which has been extensively trained on a large number of ECG-Text pairs with contrastive learning to effectively capture the intricate patterns within the ECG time series:
\begin{equation}
    \textbf{e}_{ts}\in\mathbb{R}^{n_s \times d_s} = E_{ts}(\boldsymbol{x}_{ts} | \theta_{E_{ts}}),
\end{equation}
where $n_s$ is the number of time series feature patches and $d_s$ is the dimension of time series features.
For ECG image encoder $E_{img}(\theta_{img})$, we utilize the pre-trained CLIP encoder from LLaVA \citep{liu2024visual}. This model is adept at understanding and processing visual information, making it suitable for extracting features from the ECG images:
\begin{equation}
    \textbf{e}_{img}\in\mathbb{R}^{n_m \times d_m} = E_{img}(\boldsymbol{x}_{img}  | \theta_{E_{img}}),
\end{equation}
where $n_m$ is the number of image feature patches and $d_m$ is the dimension of image features. These two encoders enable the separate extraction of features from time series and images. This dual-encoder approach allows us to harness the distinct advantages of each data type, enhancing the overall interpretative power of our model.

\subsection{Cross-modal Alignment Learning}
Considering that the time series and image encoders are trained independently, their generated representations often show inconsistencies within the representation space. Meanwhile, to ensure that the LLM can interpret ECG time series and images effectively, it is essential that these multimodal inputs are rendered as comprehensible as textual data. Therefore, aligning these diverse modality inputs within a unified representation space becomes crucial. 
 
In our approach, the ECG time series representation is first mapping to the same dimensionality as the ECG image representation. This is accomplished using a multi-layer perceptron (MLP) projector:
\begin{equation}
    \hat{\textbf{e}}_{ts}\in\mathbb{R}^{n_s \times d_m} = MLP_{ts}(\textbf{e}_{ts} | \theta_{M_{ts}}).
\end{equation}
Subsequently, we employ an additional projector to map both the time series and image representations into a consistent textual space:
\begin{equation}
    \textbf{h}_{ts}\in\mathbb{R}^{n_s \times d_t} = MLP(\hat{\textbf{e}}_{ts} | \theta_{M}), 
\end{equation}
\begin{equation}
    \textbf{h}_{img}\in\mathbb{R}^{n_m \times d_t} = MLP(\textbf{e}_{img} | \theta_{M}),
\end{equation}
where $d_t$ is the dimension of the text embeddings.
This step is for ensuring that the features extracted from both modalities are not only aligned dimensionally but are also interpretable in a LLM-friendly format. 

Once we have obtained the features from both the time series and image modalities, we integrate these with the embeddings of the textual query $\boldsymbol{x}_{q}$:
\begin{equation}
    \textbf{x} = \text{Concatenate}(\textbf{h}_{ts},\textbf{h}_{img},\text{Embeded}(\boldsymbol{x}_{q})).
\end{equation}
The integration is crucial for creating a cohesive representation that encapsulates the full spectrum of information from the multimodal inputs. 

\subsection{Knowledge-guided Instruction Data Generation}
Instruction data forms the foundation of multimodal training, directly shaping how the MLLM generates responses for given queries. This is because the language response $\theta_{LLM}(\textbf{x})$ is optimized to match the target answer $\boldsymbol{y}$ for each multimodal input $ (\boldsymbol{x}_{ts}, \boldsymbol{x}_{img}, \boldsymbol{x}_{q})$. Here, we introduce two key mechanisms to guarantee that $\boldsymbol{y}$ aligns with feature-grounded analysis, evidence-driven diagnosis, and a realistic interpretation process, thereby empowering GEM with a grounded understanding of ECG data.

\textbf{Grounding Feature Extractor.} To enable the feature-grounded analysis, we propose to further excavate more detailed information from raw ECG time series. We begin by extracting universal elements, including waveforms, amplitudes of fiducial points, and intervals, from each heartbeat in each lead. These elements are then structured into a time-ordered sequence for further analysis. For instance, an ECG time series with ten visible heartbeats allows us to construct a QRS duration sequence as \([QRS_1, QRS_2, ..., QRS_{10}]\), where each element precisely represents the QRS duration for a specific heartbeat. These sequences provide a fine grade description of the physiological features of the ECG, enabling the model to assess heart conditions at the level of individual heartbeats. In our implementation, we incorporate 14 feature sequences for each of the 12 ECG leads, covering: Heart Rate, RR Interval 1, RR Interval 2, P Amplitude, P Duration, PR Interval, QRS Amplitude, QRS Duration, T Amplitude, T Duration, ST Duration, ST Form, QT Interval, and QTc Interval. This comprehensive set of sequences captures the temporal evolution of key physiological features, enabling granular analysis of cardiac activity.
The feature extraction process is formulated as:
\begin{equation}
    \boldsymbol{x}_{fs} = \text{FeatureDB}(\boldsymbol{x}_{ts}),
\end{equation}
where $\boldsymbol{x}_{fs}$ is a dictionary in which the keys represent feature names and the values correspond to their respective feature values. $\text{FeatureDB}(\boldsymbol{x}_{ts})$ represents the function that extracts structured physiological features from the ECG time series $\boldsymbol{x}_{ts}$. Note that there is no trainable parameters in FeatureDB.

\textbf{Diagnosis Guider.} 
With the feature sequences $\boldsymbol{x}_{fs}$ extracted, the challenge lies in generating high-granularity $\boldsymbol{y}$ without relying on costly human-expert annotation. To address this, we design a diagnosis guider that constructs a prompt $\boldsymbol{x}_{p}$ to effectively guide GPT-4o in generating accurate and clinically grounded responses $\boldsymbol{y}$ for each sample:
\begin{equation}
    \boldsymbol{x}_{p} = \text{DiagnosisGuider}(\boldsymbol{x}_{fs}).
\end{equation}
The diagnosis guider provides cardiology-specific instructions for detailed analysis of each aspects of ECG data ($e.g.$, instruct GPT-4o to assess the P wave amplitude and duration in Lead II to evaluate atrial enlargement) and incorporates guidance reflecting real-world clinical diagnostic processes. As a result, each sample receives a unique $\boldsymbol{x}_{p}$ tailored to its $\boldsymbol{x}_{fs}$, ensuring precise activation of GPT-4o’s latent medical knowledge for accurate and personalized analysis.
See Appendix \ref{diag_guider} for more details of the diagnosis guider.

\begin{figure*}[t]
    \centering
    \includegraphics[width=\textwidth]{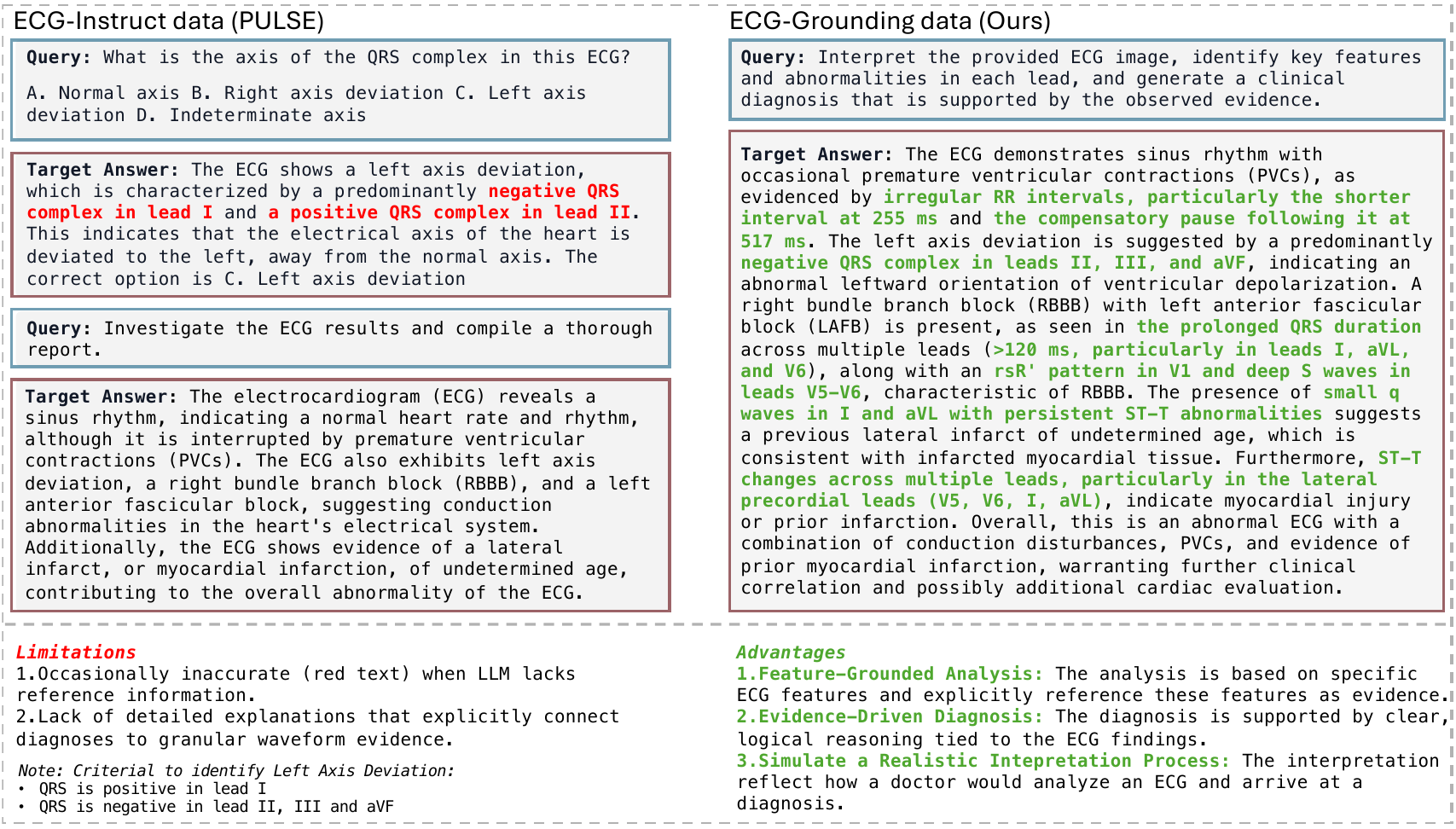}
    \caption{Comparison of ECG-Instruct and our ECG-Grounding.  %
    }
    \label{fig:instruct}
\end{figure*}

\textbf{ECG-Grounding Data.}
Using the knowledge-guided instruction data generation method, we employ GPT-4o to curate 30,000 fine-grained instruction-response pair $(\boldsymbol{x}_{q}, \boldsymbol{y})$ from the MIMIV-IV-ECG \citep{gow2023mimic} database:
\begin{equation}
    \boldsymbol{y} = \text{GPT-4o}(\boldsymbol{x}_{p}).
\end{equation}
A comparison illustrating the advantages of our instruction data is shown in Figure \ref{fig:instruct}. Instruction data from PULSE, primarily derived from the original report, occasionally provides incorrect explanations due to hallucinations, which can arise when LLM has limited reference information. In contrast, our ECG-Grounding provides more accurate, holistic, and evidence-driven interpretations with diagnoses grounded in measurable ECG features. Training with these fine-grained instructional data, GEM significantly enhances its explainability and grounding capabilities, making it a promising approach for clinical applications and fostering greater trust among medical professionals.

\subsection{Training}
\label{imple}
Unlike conventional multi-step training pipelines for most MLLMs, which first train a linear projector using brief image captions for cross-modal alignment and subsequently fine-tune the LLM with instruction data, we adopt an one-step training approach for GEM. In this approach, we freeze $\theta_{ts}$ and $\theta_{img}$ for feature extraction, while jointly training $\theta_{M_{ts}}$, $\theta_{M}$, and $\theta_{LLM}$. This one-step process enhances the consistency in training multiple modalities with limited data and improves the efficiency of the training phase. 

The training objective is formulated as minimizing the negative log-likelihood of the target answer $y$ and the LLM response $\theta_{LLM}(\textbf{x})$, given the embedding of multimodal input $\textbf{x}$:
\begin{equation}
    L(\boldsymbol{y};\theta_{LLM}(\textbf{x})) = -\sum_{i=1}^{N}logP(y_j|\textbf{x}, \theta_{LLM}),
\label{loss}
\end{equation}
where $N$ represents the number of tokens in $\boldsymbol{y}$ and $y_j$ is the $j$-th token in $\boldsymbol{y}$.
We train two foundational LLMs for the GEM framework. The first, LLaVA, is the unmodified version of the original LLaVA-7B model, which has not undergone training on any ECG data. The second, referred to as PULSE, represents state-of-the-art LLMs specifically trained on millions of ECG images. For both models, we implement supervised fine-tuning (SFT) for a single epoch. We use 8 A100 GPUs for the training.

\section{Experiments}
\label{experiment}
\subsection{Training Data}
The training of GEM involves two datasets: the ECG-Instruct data from PULSE \citep{liu2024teach}, which includes 1,156,110 conversations, and our generated ECG-Grounding data, comprising 30,000 conversations. The ECG-Grounding data is sampled from MIMIC-IV-ECG \citep{gow2023mimic}, selecting only samples that have not been used in training models such as PULSE. We use the ECG-image-kit \citep{shivashankara2024ecg} for the generation of ECG images from the original ECG signal, and use the FeatureDB \citep{hong2017encase, hong2019combining} to extract ECG features. To facilitate further research and benefit the community, we have publicly released ECG-Grounding data.

\subsection{Evaluation Tasks and Metrics}
\textbf{Grounded ECG Understanding}. To comprehensively evaluate whether the model achieves clinically grounded ECG interpretation capabilities comparable to cardiologists, we introduce the Grounded ECG Understanding task. This task is developed based on cardiologist guidelines and evaluates the MLLM’s ability to identify detailed diagnostic clues in ECG analysis, requiring it to provide specific details and relevant clinical knowledge to support its interpretation. 

We utilize GPT-4o to score the responses of the MLLM using a predefined set of metrics that measures the accuracy and comprehensiveness of the details provided. Specifically, these metrics include:
\textit{DiagnosisAccuracy}, \textit{AnalysisCompleteness}, \textit{AnalysisRelevance}, \textit{LeadAssessmentCoverage}, \textit{LeadAssessmentAccuracy}, \textit{ECGFeatureGrounding}, \textit{EvidenceBasedReasoning}, and \textit{ClinicalDiagnosticFidelity}.
The definitions and interpretations of each metric are provided in Appendix~\ref{def_eval}, and further details on the scoring methodology for each criterion are outlined in Appendix~\ref{eval_prompt}.

The evaluation is conducted on the test sets of two datasets: MIMIC-IV-ECG (2,381 samples) and PTB-XL\citep{wagner2020ptb} (2,041 samples). For the grounded ECG understanding task, MIMIC-IV-ECG serves as the in-domain dataset, while PTB-XL represents an out-domain dataset, with test samples drawn from a different data distribution.

\textbf{ECG-Bench}. In addition, we use the ECG-Bench \citep{liu2024teach} to assess our model's capability in cardiac abnormality detection and report generation. ECG-Bench is a comprehensive benchmark for evaluating MLLMs on ECG understanding. It incorporates several datasets including the PTB-XL dataset, CPSC2018 dataset \citep{liu2018open}, G12EC dataset \citep{alday2020classification}, CODE-15\% dataset \citep{ribeiro2021code}, and CSN dataset \citep{zheng202012}. ECG-Bench contains two main tasks: abnormality detection and report generation. For the abnormality detection task, it uses AUC, F1, and Hamming Loss (HL) as metrics for multi-label datasets, and accuracy for others. In the report generation task, it employs GPT-4o to evaluate the reports based on their accuracy in rhythms, waveform descriptions, and diagnoses, with a maximum score of 100.

\begin{table*}[t]
\caption{Grounded ECG Understanding results on MIMIC-IV-ECG and PTB-XL.}
\centering
\resizebox{\linewidth}{!}{
\begin{tabular}{lcccccccc}
\toprule
\textbf{Metric} & \footnotesize\textbf{\makecell[l]{Diagnosis \\ Accuracy}} & \footnotesize\textbf{\makecell[c]{Analysis \\ Completeness}} & \footnotesize\textbf{\makecell[c]{Analysis \\ Relevance}} &\footnotesize\textbf{\makecell[c]{Lead\\Assessment \\ Coverage}} &\footnotesize\textbf{\makecell[c]{Lead \\ Assessment \\ Accuracy}} &\footnotesize\textbf{\makecell[c]{ECG\\Feature\\Grounding}} &\footnotesize\textbf{\makecell[c]{Evidence\\Based \\ Reasoning}}  & \footnotesize\textbf{\makecell[c]{Clinical \\ Diagnostic \\ Fidelity}} \\
\midrule
\midrule
\multicolumn{9}{l}{\textbf{MIMIC-IV-ECG (in-domain)}} \\
PULSE & 81.14 & 2.37 & 2.39 & 7.11 & 2.95 & 50.18 & 52.40 & 51.63 \\
GEM (Ours) & \\
\quad SFT LLaVA  & \textbf{87.24} & 4.41 & \textbf{5.01} & \textbf{71.07} & \textbf{46.44} & \textbf{75.48} & \textbf{75.09} & \textbf{75.28} \\
\quad SFT PULSE  & 86.49 & \textbf{4.43} & 4.91 & 69.80 & 45.33 & 74.95 & 74.70 & 74.87\\
\midrule
\midrule
\multicolumn{9}{l}{\textbf{PTB-XL (out-domain)}} \\
PULSE & 59.24 & 2.20 & 2.06 & 11.20 & 6.27 & 52.52 & 55.48 & 53.85 \\
GEM (Ours) & \\
\quad SFT LLaVA  & 73.53 & 4.19 & 2.96 & \textbf{79.54} & \textbf{49.01} & 74.48 & 74.61 & 73.84 \\
\quad SFT PULSE  & \textbf{73.59} & \textbf{4.19} & \textbf{3.00} & 78.86 & 47.96 & \textbf{74.97} & \textbf{75.41} & \textbf{74.24}\\
\bottomrule
\end{tabular}
}
\label{tab:grounded_results_combined}
\end{table*}

\begin{table*}[t]
\centering
\caption{ECG-Bench abnormality detection results.}
\resizebox{0.9\linewidth}{!}{
\begin{tabular}{lccccccccccc}
\toprule
\textbf{Datasets} & \multicolumn{3}{c}{\textbf{PTB-XL Super}} & \multicolumn{3}{c}{\textbf{CODE-15\%}} & \multicolumn{3}{c}{\textbf{CPSC 2018}} & \textbf{CSN} & \textbf{G12EC} \\ \midrule

\textbf{Metric} & \textbf{AUC} & \textbf{F1} & \textbf{HL} & \textbf{AUC} & \textbf{F1} & \textbf{HL} & \textbf{AUC} & \textbf{F1} & \textbf{HL} & \textbf{ACC} & \textbf{ACC} \\
\midrule
Random & 50.3 & 33.2 & 50.1 & 48.8  & 15.0 & 32.1 & 51.2 & 15.1 & 28.8 & 11.6 & 12.1\\
GPT-4o & 55.6 & 28.3 & 26.2& 59.9 & 24.9 & 15.7 & 50.9 & 10.6 & 18.2 & 57.5 & 49.2\\
PULSE  & 82.4 & 74.8 & 11.0& 90.7 & 85.4 & 5.0  & 76.9 & 57.6 & 8.6 & 85.2 & 78.2 \\
\midrule
GEM (Ours) \\
\quad SFT LLaVA & 81.8 & 73.6 & 11.6& 90.5 & 84.8 & 5.1  & 74.1 & 52.0 & 9.0 & \textbf{92.6} & \textbf{81.8}  \\
\quad SFT PULSE & \textbf{83.4} & \textbf{75.8} & \textbf{11.0}& \textbf{91.5} & \textbf{86.4} & 4.7 & \textbf{79.1} & \textbf{61.1} & \textbf{8.1} & 86.2 & 80.5\\
\midrule
\midrule
Ablations \\
\quad TS only & 81.2 & 72.5 & 11.9& 90.8 & 84.9 & 5.0 & 76.3 & 54.0 & 8.5 & 91.6 & 81.4 \\ 
\quad TS+IMG & 82.7 & 74.8 & 11.1& 91.3 & 86.3 & \textbf{4.6}  & 74.4 & 51.5 & 8.8 & 90.1 & 81.1 \\
\bottomrule
\end{tabular}
}
\label{ecg_bench1}
\end{table*}

\subsection{Results}
Table~\ref{tab:grounded_results_combined} summarizes the performance of GEM on the Grounded ECG Understanding task across both in-domain (MIMIC-IV-ECG) and out-domain (PTB-XL) datasets.
The proposed GEM models consistently outperform the state-of-the-art PULSE model across all evaluation metrics.
For diagnosis accuracy, GEM achieves 87.24\% on MIMIC-IV-ECG, surpassing PULSE (81.14\%) by over 6\%. On the out-domain dataset, GEM maintains robust generalization with an accuracy of 73.59\%, again outperforming PULSE (59.24\%) by a notable margin of 14.35\%.
This remarkable performance on the out-domain dataset highlights GEM’s strong reasoning capability, enabling it to generalize effectively and make accurate diagnoses across diverse data distributions.
In analysis completeness and relevance, GEM demonstrates substantial gains. On MIMIC-IV-ECG, it improves completeness from 2.37 to 4.43 and relevance from 2.39 to above 5.01. Similar trends are observed on PTB-XL, where GEM doubles the PULSE’s completeness and improves relevance by over 45\%. This substantial performance gap indicates that GEM not only identifies and interprets a greater number of critical ECG components (with an average two-fold increase in feature coverage) but also maintains stronger clinical relevance by effectively connecting these observations to the diagnostic reasoning process. 
Lead assessment coverage and accuracy also improve significantly. On the in-domain dataset, GEM increases coverage from 7.11\% to over 71\% and accuracy from 2.95\% to above 46\%. Improvements on the out-domain dataset rising from 11.20\% to over 79\% for coverage and from 6.27\% to over 49\% for accuracy. These gains verify GEM’s ability to perform structured and precise evaluation across multiple ECG leads, an essential skill for cardiologist-level reasoning.

Regarding ECG feature grounding, GEM achieves scores around 75 on both datasets, indicating that a large portion of diagnostic conclusions are explicitly linked to measurable ECG parameters. This is a significant advancement over PULSE, which achieves scores only around 50.
Furthermore, GEM outperforms PULSE in evidence-based reasoning and clinical diagnostic fidelity, with both metrics exceeding 74 across datasets. These results show GEM’s ability to construct clinically coherent justifications and align with structured diagnostic processes, key for real-world clinical integration.

In the ECG-Bench task, Table \ref{ecg_bench1} offers a detailed examination of the models' performance on cardiac abnormality detection across a variety of datasets. 

\begin{wraptable}[11]{r}{0.55\textwidth}
\vspace{-7pt}
\centering
\caption{ECG-Bench report generation and QA results.}
\resizebox{0.9\linewidth}{!}{
\begin{tabular}{lcc}
\toprule
\textbf{Datasets} & \textbf{PTB-XL Report} & \textbf{ECG-QA} \\
\cmidrule(lr){2-3}
\textbf{Metric} &\textbf{Report Score} & \textbf{Accuracy} \\
\midrule
Random & 0  & 16.2 \\
GPT-4o & 50.2 & 35.2 \\
PULSE & 61.3 & \textbf{73.8} \\
\midrule
GEM (Ours) \\
\quad SFT LLaVA & 65.0 & 71.0\\
\quad SFT PULSE & \textbf{67.1} & 73.6 \\
\bottomrule
\end{tabular}
}
\label{ecg_bench2}
\end{wraptable}
The results clearly show that our GEM (SFT PULSE) model consistently outperforms other models. 
Furthermore, the GEM (SFT LLaVA) version, which has not been previously trained on ECG data, still manages to achieve comparable, and in 
some cases superior ($e.g.$, 7.4\% in CSN and 3.6\% in G12EC), performance across most datasets, despite being trained for only a single epoch. 
These results highlight the robustness and efficacy of the GEM framework, demonstrating its capacity to deliver substantial performance gains even with minimal domain-specific training. 
Ablation studies further highlight the critical role of multimodal inputs in achieving efficient training and superior classification performance. We assess two model variants: TS-only, which is trained exclusively on ECG time series, and TS+IMG, which incorporates both time series and image data. Notably, the TS+IMG model surpasses PULSE on PTB-XL Super and CODE-15\% datasets, despite being trained for only one epoch. In Table \ref{ecg_bench2}, GEM showcases exceptional report generation capabilities, achieving a substantial 5.8\% improvement over PULSE in the PTB-XL Report while maintaining comparable performance in ECG-QA. These results further evident GEM’s capability in delivering holistic, accurate, and clinically-aligned ECG interpretations.

Collectively, GEM demonstrates superior performance in both Grounded ECG Understanding and ECG Bench tasks, establishing its effectiveness in ECG interpretation across multiple dimensions.

\subsection{Cardiologist Evaluation}
\label{cardiologist_eval}

We conduct a structured cardiologist evaluation covering three sources of outputs: GPT-4o generated training data, Deepseek-R1 generated training data, and GEM generated interpretations.

In total, 400 ECG-Grounding data (200 from GPT-4o and 200 from Deepseek-R1 \cite{guo2025deepseek}) and 200 GEM's interpretation were independently reviewed by eight board-certified cardiologists, using seven predefined clinical criteria designed to assess both real-world reliability and usefulness. This unified evaluation protocol allows us to (1) verify the quality of GPT-4o generated training data, (2) test the effectiveness of open-source substitutes, and (3) confirm the clinical utility of the GEM model. See Appendix \ref{human_eval_metrics} for detailed scoring criteria.
 
We report the evaluation results in Table~\ref{huam_eval_reliability} and \ref{huam_eval_usefulnesss} below. The expert evaluation shows that GPT-4o consistently achieves high scores across both reliability and clinical usefulness, with particularly strong performance in analytical completeness and reasoning quality. These results demonstrate that, when using our knowledge-guided instruction data generation, GPT-4o is capable of generating high-quality ECG interpretations that are both clinically reliable and practically valuable.

Deepseek-R1 is also capable of generating clinically acceptable, high-quality ECG interpretations with our methods. This demonstrates that our method is adaptable to alternative LLM backbones and remains applicable in settings without commercial API access. 

The expert evaluation results also demonstrate that GEM consistently achieves high scores across both reliability and usefulness dimensions, with most metrics rated above 4 out of 5. 
These findings indicate that GEM is capable of producing clinically meaningful and accurate interpretations that align well with cardiologists’ expectations. 
High scores in reliability-related metrics reflect the factual correctness and clinical grounding of its outputs, while strong performance in usefulness metrics suggests that GEM is not only technically sound but also practically helpful in supporting diagnostic decision-making. 
Collectively, these results support GEM's potential as a trustworthy assistant for real-world cardiology applications.

In Appendix \ref{case_study}, we showcase six representative cases involving complex cardiac conditions, in which GEM’s interpretations exceeded expert expectations.
In these cases, cardiologists highlight two types of findings: (1) those they have not noticed during their own ECG examination, which exceed their expectations for a cardiology AI assistant in real-world settings, and (2) those where they hold differing opinions.

Overall, GEM demonstrated its capability to generate clinically insightful findings, often surpassing expert expectations by identifying details that cardiologists have not noticed, suggesting its potential for real-world clinical applications.
It is noteworthy that cardiologists also highlights certain cases where their interpretations diverged from GEM or GPT-4o.
Although our knowledge-guided instruction data generation approach avoids costly expert annotations while producing high-quality target answers, GPT-4o still occasionally generates target answers that may be misaligned with cardiologist interpretations.
For example, in Figure \ref{fig:case2}, the GPT-4o suggest no evidence ischemia or infarction, while cardiologist suspects there are ischemia in the precordial leads.
These deviations highlight opportunities for future refinement of model reasoning with human feedback to better align with cardiologist level clinical judgment.

\begin{table}[t]
\centering
\caption{Evaluation of reliability metrics by cardiologists (Mean and STD).}
\label{huam_eval_reliability}
\begin{tabular}{lccc}
\toprule
\textbf{Model} & \textbf{\makecell{Analytical\\Relevance}} & \textbf{\makecell{Analytical\\Accuracy}} & \textbf{\makecell{Analytical\\Completeness}} \\
\midrule
GPT-4o & 4.7/5 (0.66) & 4.6/5 (0.82) & 4.7/5 (0.65)\\
Deepseek-R1 & 4.8/5 (0.57) & 4.7/5 (0.78) & 4.9/5 (0.42)\\
GEM & 4.6/5 (0.60) & 4.4/5 (0.80) & 4.6/5 (0.57)\\
\bottomrule
\end{tabular}
\end{table}

\begin{table}[t]
\centering
\caption{Evaluation of usefulness metrics by cardiologists (Mean and STD).}
\label{huam_eval_usefulnesss}
\begin{tabular}{lcccc}
\toprule
\textbf{Model}&\textbf{\makecell{Reasoning\\Quality}} & \textbf{\makecell{Findings\\Novelty}} & \textbf{\makecell{Clinical\\Value}} & \textbf{\makecell{Overall\\Satisfaction}} \\
\midrule
GPT-4o & 4.7/5 (0.67)& 4.4/5 (1.18)& 4.7/5 (0.73)& 4.5/5 (0.87)\\
Deepseek-R1 & 4.8/5 (0.62)& 4.5/5 (0.91)& 4.6/5 (0.82)& 4.7/5 (0.77)\\
GEM & 4.6/5 (0.64)& 3.9/5 (1.25)& 4.3/5 (0.89)& 4.4/5 (0.82)\\
\bottomrule
\end{tabular}
\end{table}

\subsection{Failure Case Analysis}
\label{failure_case_analysis}
We also conduct an analysis of failure cases informed by expert feedback from the human evaluation. The main errors fall into two categories. The first is incorrect diagnosis, often caused by limitations in the representation stage. Subtle morphological patterns such as ST-segment changes or P-wave abnormalities were occasionally missed by the encoders, likely due to feature representation deficiencies or insufficient training data coverage of diverse ECG variations. The second is overstating the severity of findings. GEM occasionally exaggerated the severity of certain cardiac conditions, which cardiologists noted could lead to unnecessary patient concern. This tendency may result from the absence of patient context, as the model interprets ECGs without access to clinical history that physicians would normally consider. These findings help clarify the current limitations and inform future directions to enhance model safety, reliability, and clinical alignment.

\section{Conclusion}
\label{conclusion}
In this work, we present GEM, the first MLLM for grounded ECG interpretation, integrating ECG time series, 12-lead ECG images, and textual instructions. 
GEM achieves feature-grounded analysis, evidence-driven diagnosis, and clinician-style diagnostic workflows through three core technical innovations: multimodal encoding, cross-modal alignment learning, and knowledge-guided instruction data generation.
The multimodal encoding and cross-modal alignment learning allow the LLM to simultaneously process ECG time series and image representations, effectively leveraging the complementary strengths of both modalities for interpretation.
The knowledge-guided instruction data generation addresses the lack of high-granularity instruction data for ECG understanding. The developed ECG-Grounding dataset comprises 30,000 fine-grained instruction pairs annotated with heartbeat-level physiological features, which establishes the first high-resolution resource for grounded ECG understanding.
We also introduce the Grounded ECG Understanding task, a clinically motivated benchmark that comprehensively assesses models' grounded ECG understanding through multi-dimensional metrics. 
Together, these contributions establish a solid foundation for future works in conversational diagnostic AI for ECG interpretation.

\newpage

\textbf{Acknowledgment} \\
This research is supported by A*STAR, CISCO Systems (USA) Pte. Ltd., and the National University of Singapore under its Cisco-NUS Accelerated Digital Economy Corporate Laboratory (Award I21001E0002),  and the AI for Public Health Program in Saw Swee Hock School of Public Health, National University of Singapore. Shenda Hong is supported by CCF-Tencent Rhino-Bird Open Research Fund (CCF-Tencent RAGR20250108), and CCF-Zhipu Large Model Innovation Fund (CCF-Zhipu202414). 

\bibliography{ref}
\bibliographystyle{plainnat}

\newpage
\appendix

\begin{center}
    {\LARGE \textbf{GEM: Empowering MLLM for Grounded ECG Understanding with Time Series and Images \\ Appendix}}
\end{center}

\begin{appendices}

\startcontents[sections]
\printcontents[sections]{}{1}{\setcounter{tocdepth}{2}}
\newpage

\renewcommand{\thesection}{A.\arabic{section}}
\renewcommand{\thesubsection}{\thesection.\arabic{subsection}}

\section{Diagnosis Guider Prompt}
\label{diag_guider}
\begin{adjustwidth}{0cm}{0cm}
\ttfamily
\# Your task: Interpret the provided ECG image, identify key features and abnormalities in each lead, and generate a clinical diagnosis that is supported by the observed evidence.

\medskip
\#\# Key objectives:\\
    1. Simulate a Realistic Diagnostic Process: The interpretation should reflect how a doctor would analyze an ECG, ask clarifying questions, and arrive at a diagnosis.\\
    2. Grounded ECG Understanding: The analysis should be based on specific ECG features and explicitly reference these features as evidence.\\
    3. Evidence-Based Reasoning: The diagnosis should be supported by clear, logical reasoning tied to the ECG findings.

\medskip
\#\# Guidelines for the ECG analysis: \\
    1. Data:\\
        ECG image: an image that display the 12-lead ECG tracings. Make the task centered on the ECG image, assuming direct ECG image analysis.\\
        Machine measurements: A time-ordered list of ECG features computed for each heartbeat in every lead. Each entry in the list corresponds to the features calculated for a single heartbeat. 
        
\medskip
    2. Act as a cardiologist and use medical knowledge to analyze the provided ECG image step-by-step: \\
        Initial Analysis: Analyze the provided ECG image to identify key features such as rhythm, intervals, and any apparent abnormalities. \\
        Detailed Reasoning: Explain your thought process step-by-step, referencing specific ECG features (e.g., "The ST segment is elevated in leads V1-V4, which suggests anterior myocardial infarction"). \\
        Evidence-Based Diagnosis: Propose a diagnosis or differential diagnoses, justifying your conclusions with explicit ECG data.

\medskip
    3. When analyzing the ECG image, carefully analyze each lead: \\
        Lead I: Examine the QRS amplitude and duration, along with ST segment and T wave morphology. Abnormalities may indicate lateral wall issues such as left ventricular hypertrophy, bundle branch block, or lateral ischemia/infarction.\\
        Lead II: Look at the P wave amplitude and duration to assess right or left atrial enlargement; the PR interval can reveal conduction delays. ST and T wave changes here suggest inferior wall ischemia or infarction.\\
        Leads III and aVF: Primarily reflect inferior wall status. Abnormal Q waves, along with ST segment and T wave changes, point toward inferior infarction or ischemia.\\
        Lead aVL: Focuses on the high lateral region; QRS, ST, and T wave abnormalities here suggest high lateral ischemia or infarction.\\
        Lead aVR: ST elevation may indicate left main or multivessel disease, and T wave inversion can be associated with ventricular arrhythmia.\\
        Lead V1: An increased R wave, a characteristic rsR pattern, and ST-T changes help identify right ventricular hypertrophy, right bundle branch block, or ischemia.\\
        Leads V2-V4: Assessing the anterior or anteroseptal regions. The presence of Q waves, along with ST segment and T wave deviations, suggests anterior wall infarction or ischemia.\\
        Leads V5-V6: Focus on the lateral wall, where similar QRS, ST, and T wave changes can indicate lateral ischemia or infarction.

\medskip
    4. When analyzing the machine measurements, you should aware that: \\
        a. If any abnormalities appear in the computed measurements that are not mentioned in the report, you must strictly follow and trust the report. \\
        b. Evaluate and interpret the machine measurements as if you had computed them yourself. In your analysis, refer to these values as your own computed measurements rather than using phrases like "machine measurements provided".

\medskip
\#\# Guidelines for the response generation: \\
    1. Synthesize your findings to deduce a likely diagnosis or set of diagnoses. Clearly explain how the evidence supports your conclusion.\\
    2. Ensure your diagnosis is comprehensive and strictly based on the report. Do not include diagnosis that not mentioned in the report.\\ 
    3. Make sure your diagnosis are grounded in the given ECG image and machine measurements, and you should explicitly reference (e.g., specify lead and the position of the abnormal heartbeat).\\
    4. Strictly follow the output format and requirements specified in your task instructions.\\
    5. The given report only served as the ground truth for you to analyze the ECG image. The generated text must not show that you are aware of the existence of the report.\\
    6. Never make up explanations.\\

\medskip
\#\# ECG Report: \\ 
\{\{report\}\}

\medskip
\#\# ECG Machine Measurements: \\
\{\{machine\_measurements\}\}

\#\# Generation rule
The generated text must not show that you are aware of the existence of the report.
Do not include phrases like "Based the report", or "Given the ECG report".
The primary objective is to analyze the ECG and identify evidence that supports the results. The analysis should focus solely on the ECG itself, never analyze the report.

\#\# Present your work in this format:

**Response:** [Comprehensive response following the task's guidelines, strictly based on the report. Using a complete paragraph with more natural expression. Do not use a list format. Limit your responses within 300 words.]

\end{adjustwidth}

\section{Grounded ECG Understanding Metrics}
\label{def_eval}
\textit{DiagnosisAccuracy} evaluates whether the generated diagnosis is correct, specific, and supported by ECG findings. Results are expressed as a percentage, indicating the average accuracy across identified key diagnoses. 

\textit{AnalysisCompleteness} checks if all key ECG components ($e.g.$, rhythm, intervals, waveforms, and lead-specific findings) are discussed. Results are provided in absolute terms, indicating the average number of correctly addressed key ECG features for each sample.
 
\textit{AnalysisRelevance} assesses whether each explanation directly supports the diagnosis, with results showing on average how many points support the diagnosis with clear ECG evidence for each sample.
 
\textit{LeadAssessmentCoverage} evaluates how many of the 12 ECG leads are analyzed. Results indicate the average percentage of leads analyzed per sample, providing insight into the comprehensiveness of the ECG assessment.
 
\textit{LeadAssessmentAccuracy} verifies the accuracy of described lead findings ($e.g.$, QRS, ST, T waves, amplitude, intervals, ST segments) against the ground truth interpretation. The result reflects the average percentage of accurately identified findings across the 12 ECG leads. 

\textit{ECGFeatureGrounding} determines if the interpretation references actual ECG features ($e.g.$, QRS amplitude, PR interval) instead of generic terms. Results are scaled from 0 to 100.
 
\textit{EvidenceBasedReasoning} evaluates whether the diagnosis follows logical, evidence-supported steps. Results range from 0 to 100.
 
\textit{ClinicalDiagnosticFidelity} assesses if the model mimics how a clinician interprets ECG data, considering all relevant factors. Results are scaled from 0 to 100.

\section{GPT-4o Evaluation Prompt}
\label{eval_prompt}
\begin{adjustwidth}{0cm}{0cm}
\ttfamily

\# Your task: Evaluate the alignment and quality of a generated ECG interpretation by comparing it to a ground truth clinician’s interpretation.

\#\# Evaluation Criteria:

    1. DiagnosisAccuracy: Evaluates whether the generated diagnosis is correct, specific, and supported by ECG findings. \\
        - Scoring \\
            +2 per diagnosis: Each correctly identified key diagnosis with supporting ECG features. \\
            +1 per diagnosis: Each mostly correct diagnosis but lacking key supporting details. \\
            +0 per diagnosis: Each incorrect or vague diagnosis not supported by ECG features. 

    2. AnalysisCompleteness: Checks if all key ECG components (rhythm, intervals, waveforms, and lead-specific findings) are discussed. \\
        - Scoring \\
            +1 per feature: For each correctly addressed key ECG feature (e.g., rhythm, PR interval, QRS duration, ST segment, T wave morphology). \\ 
            +0 per missing feature: For each key feature omitted or inaccurately described. 

    3. AnalysisRelevance: Assesses whether each provided explanation directly supports the diagnosis. \\
        - Scoring \\
            +2 per feature or per lead: Each point that strongly supports the diagnosis with clear ECG evidence. \\
            +1 per feature or per lead: Some points are relevant but not fully justified.\\
            +0: Includes unrelated or misleading explanations.
            
    4. LeadAssessmentCoverage: Evaluates how many of the 12 ECG leads are analyzed.\\
        - Scoring\\
            +1 per lead: For each lead adequately assessed.\\
            +0 per missing lead: For each lead omitted or inaccurately described.

    5. LeadAssessmentAccuracy: Checks if the described lead findings (e.g., QRS, ST, T waves, amplitude, intervals, ST segments) match standard ECG interpretation.\\
        - Scoring \\
            +2 per lead: Findings closely match expected values.\\
            +1 per lead: Findings are somewhat accurate but have minor inconsistencies.\\
            +0 per lead: Findings contradict ECG norms.

    6. ECGFeatureGrounding: Determines if the interpretation references actual ECG features (e.g., QRS amplitude, PR interval) instead of generic terms.\\
        - Scoring (0-100)\\
            100: ECG findings are comprehensively cited, linked to diagnoses, and cover all relevant ECG features.\\
            80: ECG findings are explicitly cited and linked to diagnoses.\\
            50: Some ECG references exist but are incomplete.\\
            0: Lacks specific waveform references.

    7. EvidenceBasedReasoning: Evaluates whether the diagnosis follows logical, evidence-supported steps.\\
        - Scoring (0-100)\\
            100: Findings logically progress to diagnosis with thorough and clear justifications covering all necessary steps.\\
            80: Findings logically progress to diagnosis with clear justifications.\\
            50: Some reasoning exists but lacks complete step-by-step analysis.\\
            0: Reasoning is unclear or not derived from ECG findings.

    8. ClinicalDiagnosticFidelity: Assesses if the model mimics how a clinician interprets an ECG, considering all relevant factors.\\
        - Scoring (0-100)\\
            100: The analysis follows a structured clinical approach and considers all relevant clinical factors.\\
            80: The analysis follows a structured clinical approach.\\
            50: Some clinical reasoning is present but incomplete.\\
            0: The approach lacks structured clinical reasoning.

    NOTE: Each score must be calculated based on strict criteria to ensure objective evaluation.

\#\# Generated ECG Interpretation: \\
\{\{model\_generated\}\} 

\#\# Ground Truth Clinician’s Interpretation:\\
\{\{groundtruth\}\}

\#\# Response Format:

Provide your evaluation strictly in the JSON format below. For any criterion with multiple elements (e.g., multiple diagnoses or leads), list each one as a separate {"Score": X, "Explanation": "Y"} entry. Use a single entry for criteria with aggregate scores (e.g., 0–100 scores).

\{
  "DiagnosisAccuracy": [
    {"Score": 2, "Explanation": "Sinus tachycardia correctly identified and supported by short PR interval."},
    {"Score": 1, "Explanation": "Left ventricular hypertrophy is mostly correct but lacks QRS amplitude detail."}
  ],
  
  "AnalysisCompleteness": [
    {"Score": 1, "Explanation": "PR interval is correctly described."},
    {"Score": 1, "Explanation": "QRS duration assessed."},
    {"Score": 0, "Explanation": "ST segment not addressed."}
  ],
  
  "AnalysisRelevance": [
    {"Score": 2, "Explanation": "QRS prolongation supports diagnosis of bundle branch block."}
  ],
  
  "LeadAssessmentCoverage": [
    {"Score": 1, "Explanation": "Lead I assessed."},
    {"Score": 1, "Explanation": "Lead II assessed."},
    {"Score": 0, "Explanation": "Leads V4–V6 omitted."}
  ],
  
  "LeadAssessmentAccuracy": [
    {"Score": 2, "Explanation": "Findings in Lead II match standard interpretation."},
    {"Score": 1, "Explanation": "Lead III slightly misinterpreted but largely accurate."}
  ],
  
  "ECGFeatureGrounding": [
    {"Score": 80, "Explanation": "Most findings cite ECG features like QRS and T wave, but some are vague."}
  ],
  
  "EvidenceBasedReasoning": [
    {"Score": 100, "Explanation": "Diagnosis is built on step-wise reasoning with reference to all major findings."}
  ],
  
  "ClinicalDiagnosticFidelity": [
    {"Score": 80, "Explanation": "Analysis mimics clinician structure but misses minor clinical context."}
  ]
\}

\end{adjustwidth}

\section{Scoring Criteria for Cardiologist Evaluation}
\label{human_eval_metrics}
\begin{longtable}{|p{2.5cm}|p{10.5cm}|}
\caption{Reliability metrics.}
\label{tab:human_eval_reliability}\\
\hline
\textbf{Criterion} & \textbf{Description and Scale} \\
\hline
\textbf{\makecell{Analytical\\Relevance\\(1–5)}} & Does the model’s analysis closely support the diagnosis and provide corresponding ECG evidence?  
\newline
5 – Every analysis point is highly relevant to the diagnosis, with clear supporting evidence.  
\newline
4 – Most analyses are strongly relevant, with minor insufficiencies.  
\newline
3 – Some analyses are relevant, but there is clear irrelevant content.  
\newline
2 – Most analyses are weakly relevant.  
\newline
1 – The analysis is unrelated to the diagnosis. 
\\ \hline

\textbf{\makecell{Analytical\\Accuracy\\(1--5)}} & 
Are there any medical factual errors in the model’s output?  
\newline
5 – Completely accurate.  
\newline
4 – Mostly accurate.  
\newline
3 – Some errors.  
\newline
2 – Obvious errors.  
\newline
1 – Severe errors. 
\\ \hline

\textbf{\makecell{Analytical\\Completeness\\(1–5)}} & 
Does the model comprehensively discuss key ECG components relevant to the diagnosis, including rhythm, intervals, and waveforms?  
\newline
5 – All relevant ECG features (rhythm, PR, QRS, ST, T waves, intervals, etc.) are accurately discussed.  
\newline
4 – Most key ECG features are covered, with minor omissions.  
\newline
3 – Only some features are covered, with significant gaps.  
\newline
2 – Only a few ECG features are mentioned.  
\newline
1 – ECG components are largely missing, with severe omissions. 
\\ \hline
\end{longtable}


\newpage

\begin{longtable}{|p{2.5cm}|p{10.5cm}|}
\caption{Usefulness metrics.} 
\label{tab:human_eval_usefulness} \\
\hline
\textbf{Criterion} & \textbf{Description and Scale} \\
\hline

\textbf{\makecell{Reasoning\\Quality\\(1–5)}} & 
Does the model provide a clear, evidence-based reasoning process similar to that of a clinician, logically deriving the diagnosis from ECG features?
\newline
5 – Clear and coherent reasoning structure, explaining each step from ECG to diagnosis causally.  
\newline
4 – Overall reasonable reasoning, but some steps lack detail.  
\newline
3 – Partial reasoning present, but incomplete or logically weak.  
\newline
2 – Disjointed reasoning with major gaps.  
\newline
1 – No logical reasoning, only a stack of conclusions. 
\\ \hline

\textbf{\makecell{Findings\\Novelty\\(1–5)}} & 
Does the model provide insights or findings not noticed by the clinician?  
\newline
5 – Important new diagnoses or findings.  
\newline
4 –  Novel and somewhat insightful content.  
\newline
3 – Some new findings, but of limited value.  
\newline
2 – Conventional content, not particularly insightful.  
\newline
1 – No new information. 
\\ \hline

\textbf{\makecell{Clinical\\Value\\(1–5)}} & 
Does the model output help in clinical decision-making?  
\newline
5 – Direct and significant support for clinical judgment; content is clear and reliable.  
\newline
4 – Most content is helpful and practically useful.  
\newline
3 – Somewhat informative, but basic or unclear.  
\newline
2 – Partially suggestive, with limited decision support.  
\newline
1 – No value for clinical judgment; not informative. 
\\ \hline

\textbf{\makecell{Overall\\Satisfaction\\(1–5)}} & 
Subjective rating of the overall output quality.  
\newline
5 – Very satisfied.  
\newline
4 – Satisfied.  
\newline
3 – Neutral.  
\newline
2 – Dissatisfied.  
\newline
1 – Very dissatisfied. 
\\ \hline
\end{longtable}

\clearpage
\section{Case Studies by Cardiologist}
\label{case_study}
%
\begin{figure}[!h]
    \centering
    \includegraphics[width=0.8\linewidth]{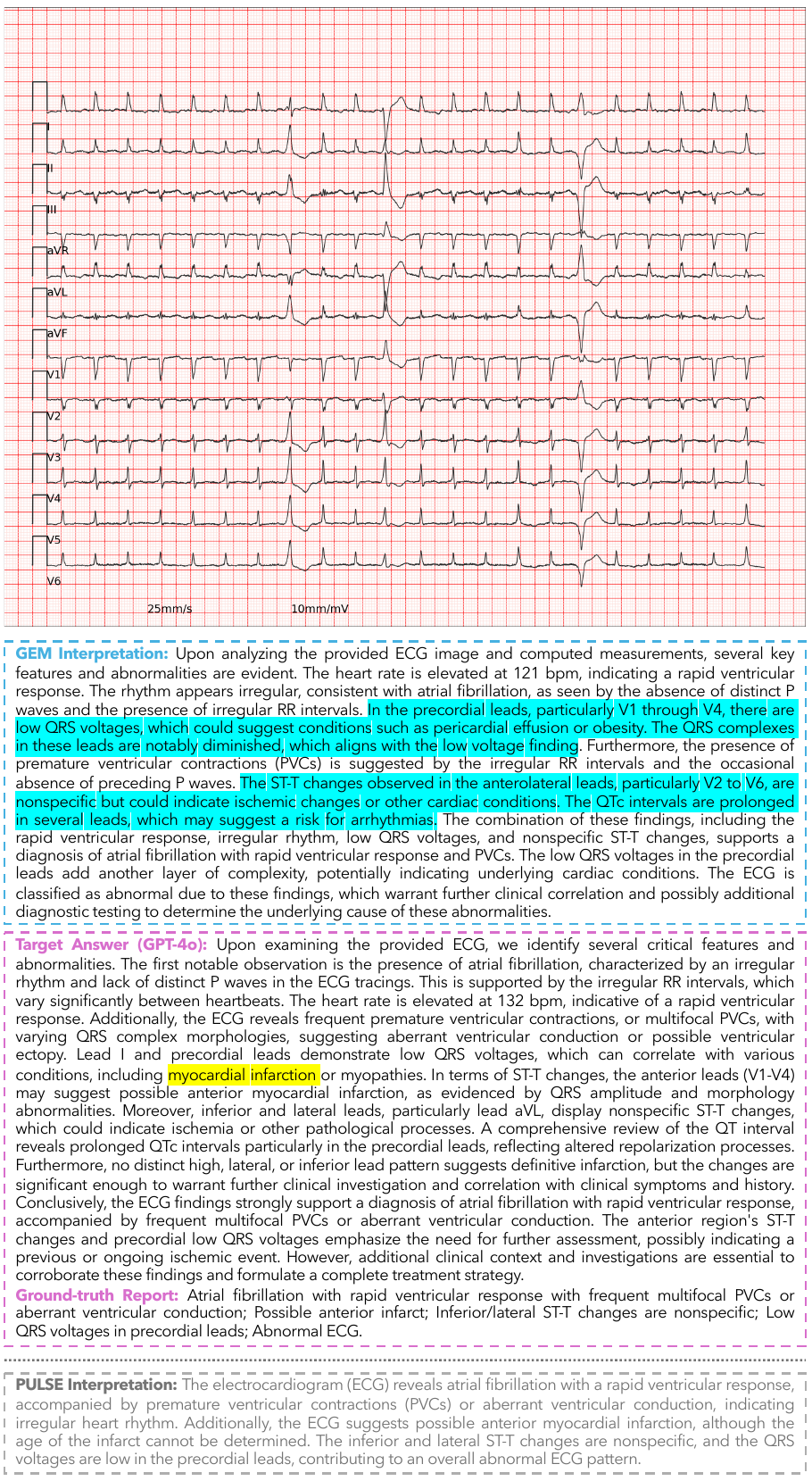}
    \caption{Cardiologist Evaluations. Blue: Findings exceeding expert expectations. Yellow: Findings with differing expert opinions.}
    \label{fig:case1}
\end{figure}

\begin{figure}[!h]
    \centering
    \includegraphics[width=0.85\linewidth]{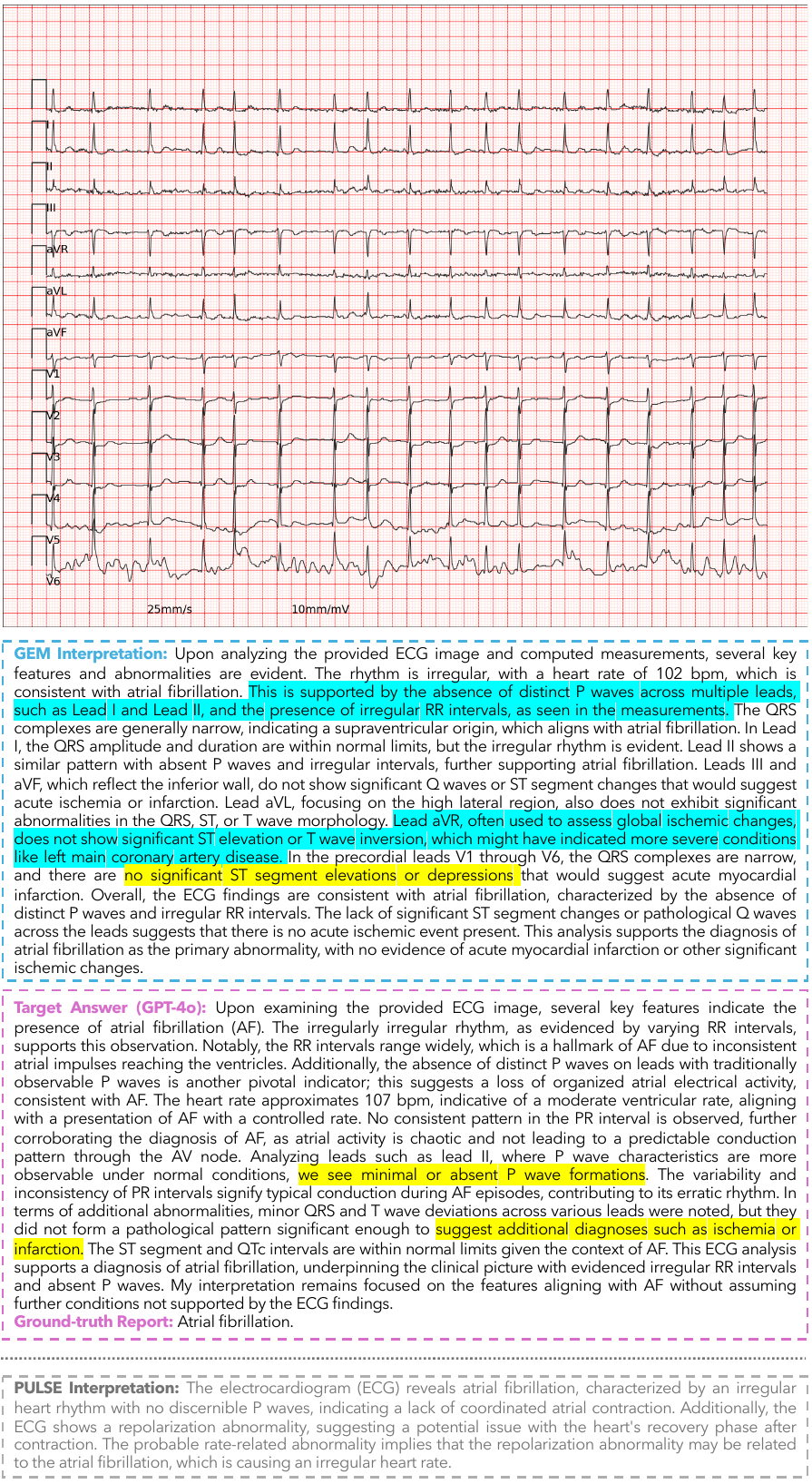}
    \caption{Cardiologist Evaluations. Blue: Findings exceeding expert expectations. Yellow: Findings with differing expert opinions.}
    \label{fig:case2}
\end{figure}

\begin{figure}[!h]
    \centering
    \includegraphics[width=0.85\linewidth]{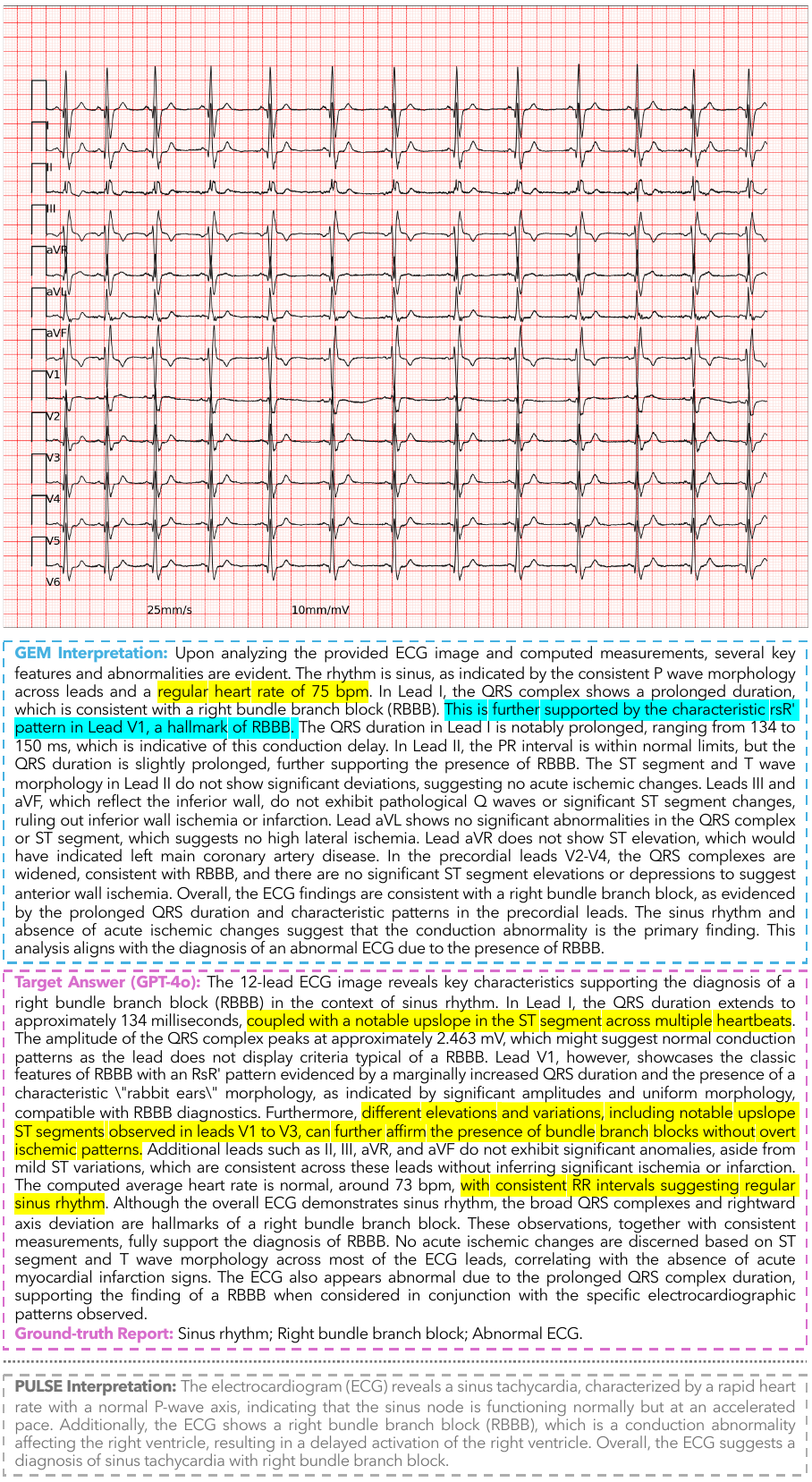}
    \caption{Cardiologist Evaluations. Blue: Findings exceeding expert expectations. Yellow: Findings with differing expert opinions.}
    \label{fig:case3}
\end{figure}

\begin{figure}[!h]
    \centering
    \includegraphics[width=0.85\linewidth]{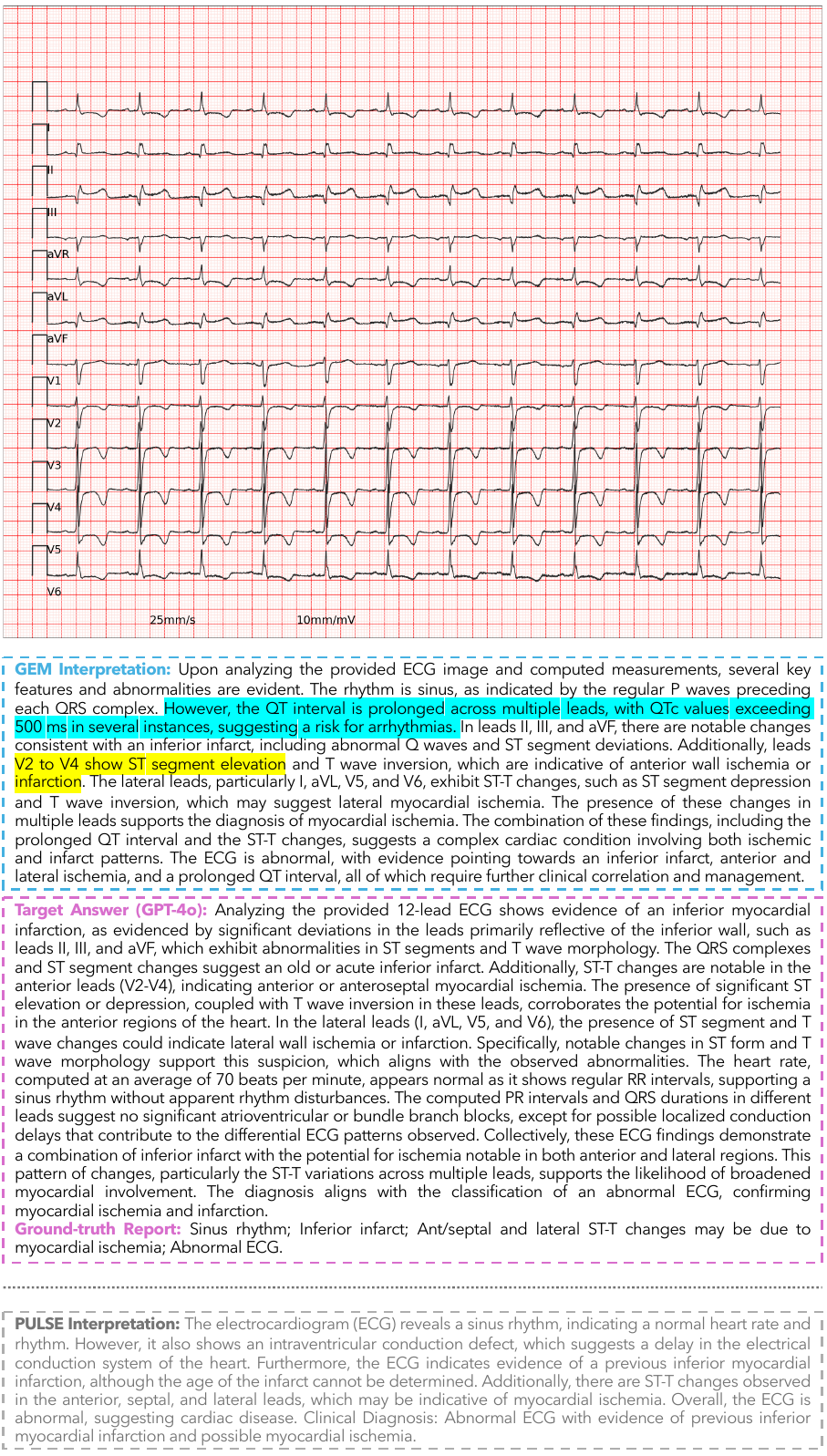}
    \caption{Cardiologist Evaluations. Blue: Findings exceeding expert expectations. Yellow: Findings with differing expert opinions.}
    \label{fig:case4}
\end{figure}

\begin{figure}[!h]
    \centering
    \includegraphics[width=0.85\linewidth]{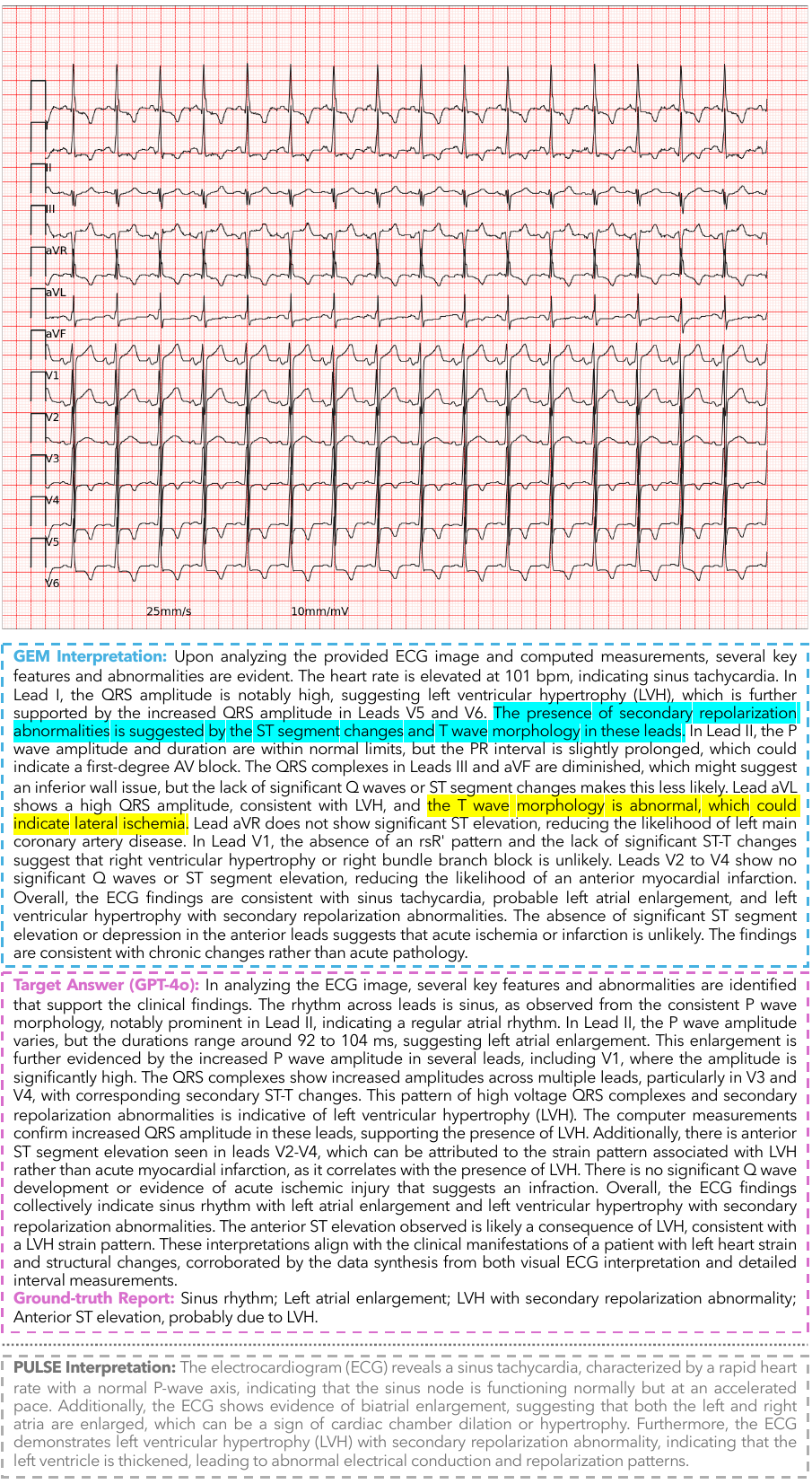}
    \caption{Cardiologist Evaluations. Blue: Findings exceeding expert expectations. Yellow: Findings with differing expert opinions.}
    \label{fig:case5}
\end{figure}

\begin{figure}[!h]
    \centering
    \includegraphics[width=0.85\linewidth]{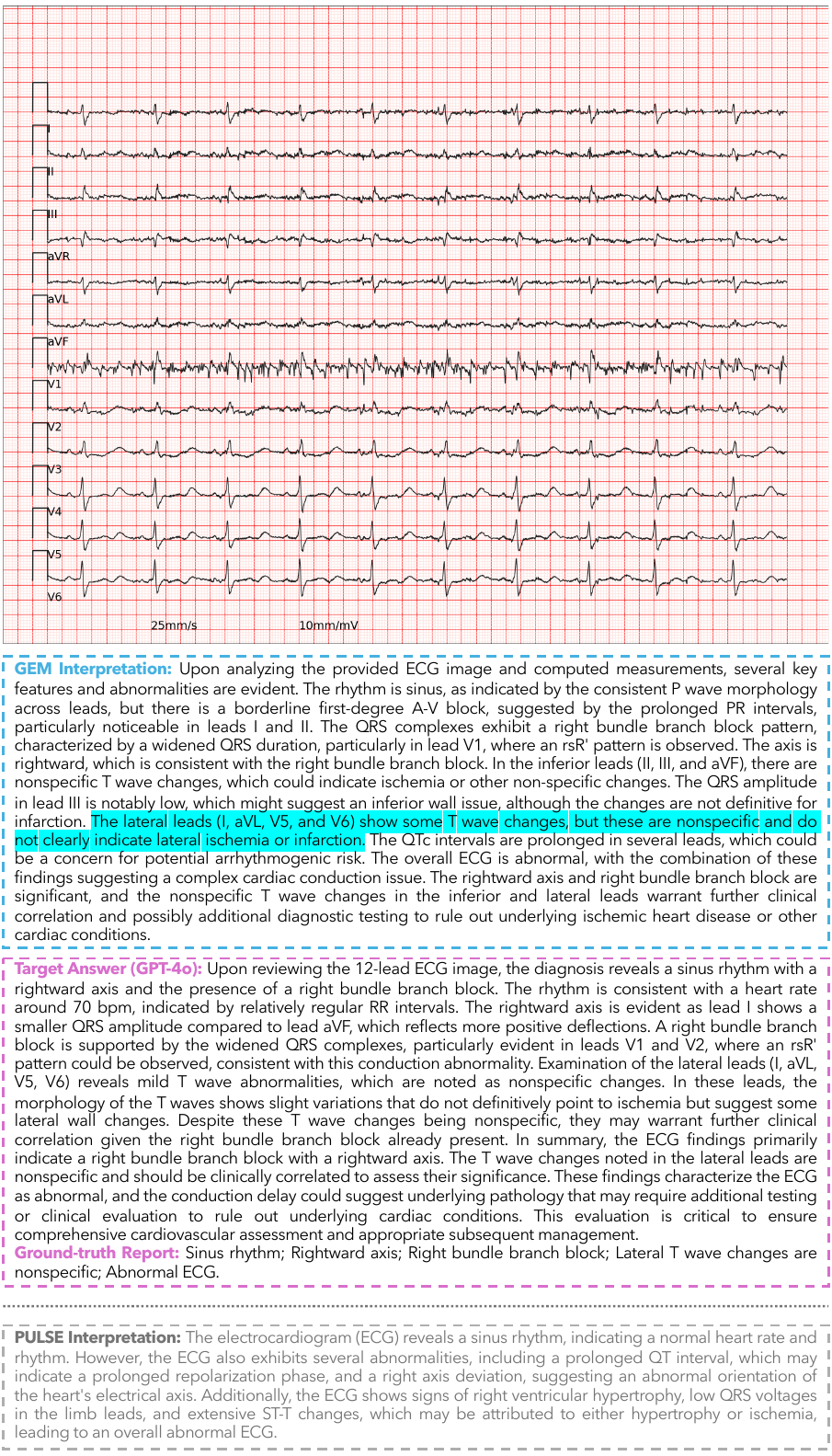}
    \caption{Cardiologist Evaluations. Blue: Findings exceeding expert expectations. Yellow: Findings with differing expert opinions.}
    \label{fig:case6}
\end{figure}
\newpage

\clearpage
\section{Broader Impacts and Limitations}
\label{braoder_impacts}
\textbf{Broader Impacts.}
This work explores the potential of leveraging large language models to generate high-granularity ECG interpretations without the need for manual annotation.
It contributes to the medical AI community in three key aspects:
First, GEM is the first multimodal large language model that unifies time series, images, and text to enable feature-grounded analysis, evidence-based diagnosis, and clinician-like diagnostic workflows. This opens new avenues for research on multimodal alignment in ECG interpretation.
Second, the proposed data generation methodology and the resulting ECG-Grounding dataset establish the first high-resolution resource for grounded ECG understanding. This dataset can support future research aimed at developing clinically applicable conversational ECG models.
Third, the introduction of the Grounded ECG Understanding task provides a comprehensive and fine-grained evaluation protocol, helping future models to be assessed more thoroughly and precisely in terms of clinical reasoning and interpretability.
By bridging computational precision with clinician-level reasoning, GEM represents a step toward more reliable, explainable, and clinically applicable AI-assisted ECG analysis, with potential for broader integration into real-world healthcare workflows.

On the other hand, it is important to acknowledge that the techniques explored in this study are intended for research purposes. Therefore, the proposed GEM model should not be directly used to make critical clinical decisions. Despite its strong performance in ECG interpretation, GEM is designed to serve as an assistive tool rather than a standalone solution for high-stakes clinical use. 

\textbf{Limitations.}
As discussed in Section~\ref{cardiologist_eval}, although our knowledge-guided instruction data generation approach avoids costly expert annotations while producing high-quality target answers, GPT-4o still occasionally generates responses that may not fully align with cardiologist interpretations. This limitation could be addressed by incorporating expert feedback into the data generation loop, which we identify as an important direction for future work.

\end{appendices}

\newpage
\section*{NeurIPS Paper Checklist}

\begin{enumerate}

\item {\bf Claims}
    \item[] Question: Do the main claims made in the abstract and introduction accurately reflect the paper's contributions and scope?
    \item[] Answer: \answerYes{} 
    \item[] Justification: The main claims made in the abstract and introduction accurately reflect the paper's contributions and scope.
    \item[] Guidelines:
    \begin{itemize}
        \item The answer NA means that the abstract and introduction do not include the claims made in the paper.
        \item The abstract and/or introduction should clearly state the claims made, including the contributions made in the paper and important assumptions and limitations. A No or NA answer to this question will not be perceived well by the reviewers. 
        \item The claims made should match theoretical and experimental results, and reflect how much the results can be expected to generalize to other settings. 
        \item It is fine to include aspirational goals as motivation as long as it is clear that these goals are not attained by the paper. 
    \end{itemize}

\item {\bf Limitations}
    \item[] Question: Does the paper discuss the limitations of the work performed by the authors?
    \item[] Answer: \answerYes{} 
    \item[] Justification: The limitations are discussed in Section \ref{braoder_impacts}.
    \item[] Guidelines:
    \begin{itemize}
        \item The answer NA means that the paper has no limitation while the answer No means that the paper has limitations, but those are not discussed in the paper. 
        \item The authors are encouraged to create a separate "Limitations" section in their paper.
        \item The paper should point out any strong assumptions and how robust the results are to violations of these assumptions (e.g., independence assumptions, noiseless settings, model well-specification, asymptotic approximations only holding locally). The authors should reflect on how these assumptions might be violated in practice and what the implications would be.
        \item The authors should reflect on the scope of the claims made, e.g., if the approach was only tested on a few datasets or with a few runs. In general, empirical results often depend on implicit assumptions, which should be articulated.
        \item The authors should reflect on the factors that influence the performance of the approach. For example, a facial recognition algorithm may perform poorly when image resolution is low or images are taken in low lighting. Or a speech-to-text system might not be used reliably to provide closed captions for online lectures because it fails to handle technical jargon.
        \item The authors should discuss the computational efficiency of the proposed algorithms and how they scale with dataset size.
        \item If applicable, the authors should discuss possible limitations of their approach to address problems of privacy and fairness.
        \item While the authors might fear that complete honesty about limitations might be used by reviewers as grounds for rejection, a worse outcome might be that reviewers discover limitations that aren't acknowledged in the paper. The authors should use their best judgment and recognize that individual actions in favor of transparency play an important role in developing norms that preserve the integrity of the community. Reviewers will be specifically instructed to not penalize honesty concerning limitations.
    \end{itemize}

\item {\bf Theory assumptions and proofs}
    \item[] Question: For each theoretical result, does the paper provide the full set of assumptions and a complete (and correct) proof?
    \item[] Answer: \answerNA{} 
    \item[] Justification: The paper does not include theoretical results.
    \item[] Guidelines:
    \begin{itemize}
        \item The answer NA means that the paper does not include theoretical results. 
        \item All the theorems, formulas, and proofs in the paper should be numbered and cross-referenced.
        \item All assumptions should be clearly stated or referenced in the statement of any theorems.
        \item The proofs can either appear in the main paper or the supplemental material, but if they appear in the supplemental material, the authors are encouraged to provide a short proof sketch to provide intuition. 
        \item Inversely, any informal proof provided in the core of the paper should be complemented by formal proofs provided in appendix or supplemental material.
        \item Theorems and Lemmas that the proof relies upon should be properly referenced. 
    \end{itemize}

    \item {\bf Experimental result reproducibility}
    \item[] Question: Does the paper fully disclose all the information needed to reproduce the main experimental results of the paper to the extent that it affects the main claims and/or conclusions of the paper (regardless of whether the code and data are provided or not)?
    \item[] Answer: \answerYes{} 
    \item[] Justification: Source codes are provided with instructions.
    \item[] Guidelines:
    \begin{itemize}
        \item The answer NA means that the paper does not include experiments.
        \item If the paper includes experiments, a No answer to this question will not be perceived well by the reviewers: Making the paper reproducible is important, regardless of whether the code and data are provided or not.
        \item If the contribution is a dataset and/or model, the authors should describe the steps taken to make their results reproducible or verifiable. 
        \item Depending on the contribution, reproducibility can be accomplished in various ways. For example, if the contribution is a novel architecture, describing the architecture fully might suffice, or if the contribution is a specific model and empirical evaluation, it may be necessary to either make it possible for others to replicate the model with the same dataset, or provide access to the model. In general. releasing code and data is often one good way to accomplish this, but reproducibility can also be provided via detailed instructions for how to replicate the results, access to a hosted model (e.g., in the case of a large language model), releasing of a model checkpoint, or other means that are appropriate to the research performed.
        \item While NeurIPS does not require releasing code, the conference does require all submissions to provide some reasonable avenue for reproducibility, which may depend on the nature of the contribution. For example
        \begin{enumerate}
            \item If the contribution is primarily a new algorithm, the paper should make it clear how to reproduce that algorithm.
            \item If the contribution is primarily a new model architecture, the paper should describe the architecture clearly and fully.
            \item If the contribution is a new model (e.g., a large language model), then there should either be a way to access this model for reproducing the results or a way to reproduce the model (e.g., with an open-source dataset or instructions for how to construct the dataset).
            \item We recognize that reproducibility may be tricky in some cases, in which case authors are welcome to describe the particular way they provide for reproducibility. In the case of closed-source models, it may be that access to the model is limited in some way (e.g., to registered users), but it should be possible for other researchers to have some path to reproducing or verifying the results.
        \end{enumerate}
    \end{itemize}

\item {\bf Open access to data and code}
    \item[] Question: Does the paper provide open access to the data and code, with sufficient instructions to faithfully reproduce the main experimental results, as described in supplemental material?
    \item[] Answer: \answerYes{} 
    \item[] Justification: Source codes are provided with instructions.
    \item[] Guidelines:
    \begin{itemize}
        \item The answer NA means that paper does not include experiments requiring code.
        \item Please see the NeurIPS code and data submission guidelines (\url{https://nips.cc/public/guides/CodeSubmissionPolicy}) for more details.
        \item While we encourage the release of code and data, we understand that this might not be possible, so “No” is an acceptable answer. Papers cannot be rejected simply for not including code, unless this is central to the contribution (e.g., for a new open-source benchmark).
        \item The instructions should contain the exact command and environment needed to run to reproduce the results. See the NeurIPS code and data submission guidelines (\url{https://nips.cc/public/guides/CodeSubmissionPolicy}) for more details.
        \item The authors should provide instructions on data access and preparation, including how to access the raw data, preprocessed data, intermediate data, and generated data, etc.
        \item The authors should provide scripts to reproduce all experimental results for the new proposed method and baselines. If only a subset of experiments are reproducible, they should state which ones are omitted from the script and why.
        \item At submission time, to preserve anonymity, the authors should release anonymized versions (if applicable).
        \item Providing as much information as possible in supplemental material (appended to the paper) is recommended, but including URLs to data and code is permitted.
    \end{itemize}

\item {\bf Experimental setting/details}
    \item[] Question: Does the paper specify all the training and test details (e.g., data splits, hyperparameters, how they were chosen, type of optimizer, etc.) necessary to understand the results?
    \item[] Answer: \answerYes{} 
    \item[] Justification: The full details are provided with the code.
    \item[] Guidelines:
    \begin{itemize}
        \item The answer NA means that the paper does not include experiments.
        \item The experimental setting should be presented in the core of the paper to a level of detail that is necessary to appreciate the results and make sense of them.
        \item The full details can be provided either with the code, in appendix, or as supplemental material.
    \end{itemize}

\item {\bf Experiment statistical significance}
    \item[] Question: Does the paper report error bars suitably and correctly defined or other appropriate information about the statistical significance of the experiments?
    \item[] Answer: \answerNo{} 
    \item[] Justification: We have conducted a comprehensive evaluation of our proposed methods across a wide range of tasks and models. The results consistently demonstrate the effectiveness of our approach. Given the breadth and robustness of these findings, we have chosen not to include formal significance testing.
    \item[] Guidelines:
    \begin{itemize}
        \item The answer NA means that the paper does not include experiments.
        \item The authors should answer "Yes" if the results are accompanied by error bars, confidence intervals, or statistical significance tests, at least for the experiments that support the main claims of the paper.
        \item The factors of variability that the error bars are capturing should be clearly stated (for example, train/test split, initialization, random drawing of some parameter, or overall run with given experimental conditions).
        \item The method for calculating the error bars should be explained (closed form formula, call to a library function, bootstrap, etc.)
        \item The assumptions made should be given (e.g., Normally distributed errors).
        \item It should be clear whether the error bar is the standard deviation or the standard error of the mean.
        \item It is OK to report 1-sigma error bars, but one should state it. The authors should preferably report a 2-sigma error bar than state that they have a 96\% CI, if the hypothesis of Normality of errors is not verified.
        \item For asymmetric distributions, the authors should be careful not to show in tables or figures symmetric error bars that would yield results that are out of range (e.g. negative error rates).
        \item If error bars are reported in tables or plots, The authors should explain in the text how they were calculated and reference the corresponding figures or tables in the text.
    \end{itemize}

\item {\bf Experiments compute resources}
    \item[] Question: For each experiment, does the paper provide sufficient information on the computer resources (type of compute workers, memory, time of execution) needed to reproduce the experiments?
    \item[] Answer: \answerYes{} 
    \item[] Justification: We provide these information in Section \ref{imple}.
    \item[] Guidelines:
    \begin{itemize}
        \item The answer NA means that the paper does not include experiments.
        \item The paper should indicate the type of compute workers CPU or GPU, internal cluster, or cloud provider, including relevant memory and storage.
        \item The paper should provide the amount of compute required for each of the individual experimental runs as well as estimate the total compute. 
        \item The paper should disclose whether the full research project required more compute than the experiments reported in the paper (e.g., preliminary or failed experiments that didn't make it into the paper). 
    \end{itemize}
    
\item {\bf Code of ethics}
    \item[] Question: Does the research conducted in the paper conform, in every respect, with the NeurIPS Code of Ethics \url{https://neurips.cc/public/EthicsGuidelines}?
    \item[] Answer: \answerYes{} 
    \item[] Justification: This paper adheres to the NeurIPS Code of Ethics.
    \item[] Guidelines:
    \begin{itemize}
        \item The answer NA means that the authors have not reviewed the NeurIPS Code of Ethics.
        \item If the authors answer No, they should explain the special circumstances that require a deviation from the Code of Ethics.
        \item The authors should make sure to preserve anonymity (e.g., if there is a special consideration due to laws or regulations in their jurisdiction).
    \end{itemize}

\item {\bf Broader impacts}
    \item[] Question: Does the paper discuss both potential positive societal impacts and negative societal impacts of the work performed?
    \item[] Answer: \answerYes{} 
    \item[] Justification: The broader impacts are discussed in Section \ref{braoder_impacts}.
    \item[] Guidelines:
    \begin{itemize}
        \item The answer NA means that there is no societal impact of the work performed.
        \item If the authors answer NA or No, they should explain why their work has no societal impact or why the paper does not address societal impact.
        \item Examples of negative societal impacts include potential malicious or unintended uses (e.g., disinformation, generating fake profiles, surveillance), fairness considerations (e.g., deployment of technologies that could make decisions that unfairly impact specific groups), privacy considerations, and security considerations.
        \item The conference expects that many papers will be foundational research and not tied to particular applications, let alone deployments. However, if there is a direct path to any negative applications, the authors should point it out. For example, it is legitimate to point out that an improvement in the quality of generative models could be used to generate deepfakes for disinformation. On the other hand, it is not needed to point out that a generic algorithm for optimizing neural networks could enable people to train models that generate Deepfakes faster.
        \item The authors should consider possible harms that could arise when the technology is being used as intended and functioning correctly, harms that could arise when the technology is being used as intended but gives incorrect results, and harms following from (intentional or unintentional) misuse of the technology.
        \item If there are negative societal impacts, the authors could also discuss possible mitigation strategies (e.g., gated release of models, providing defenses in addition to attacks, mechanisms for monitoring misuse, mechanisms to monitor how a system learns from feedback over time, improving the efficiency and accessibility of ML).
    \end{itemize}
    
\item {\bf Safeguards}
    \item[] Question: Does the paper describe safeguards that have been put in place for responsible release of data or models that have a high risk for misuse (e.g., pretrained language models, image generators, or scraped datasets)?
    \item[] Answer: \answerNA{} 
    \item[] Justification: The paper poses no such risks.
    \item[] Guidelines:
    \begin{itemize}
        \item The answer NA means that the paper poses no such risks.
        \item Released models that have a high risk for misuse or dual-use should be released with necessary safeguards to allow for controlled use of the model, for example by requiring that users adhere to usage guidelines or restrictions to access the model or implementing safety filters. 
        \item Datasets that have been scraped from the Internet could pose safety risks. The authors should describe how they avoided releasing unsafe images.
        \item We recognize that providing effective safeguards is challenging, and many papers do not require this, but we encourage authors to take this into account and make a best faith effort.
    \end{itemize}

\item {\bf Licenses for existing assets}
    \item[] Question: Are the creators or original owners of assets (e.g., code, data, models), used in the paper, properly credited and are the license and terms of use explicitly mentioned and properly respected?
    \item[] Answer: \answerYes{} 
    \item[] Justification: All assets are publicly available and properly cited in the paper.
    \item[] Guidelines:
    \begin{itemize}
        \item The answer NA means that the paper does not use existing assets.
        \item The authors should cite the original paper that produced the code package or dataset.
        \item The authors should state which version of the asset is used and, if possible, include a URL.
        \item The name of the license (e.g., CC-BY 4.0) should be included for each asset.
        \item For scraped data from a particular source (e.g., website), the copyright and terms of service of that source should be provided.
        \item If assets are released, the license, copyright information, and terms of use in the package should be provided. For popular datasets, \url{paperswithcode.com/datasets} has curated licenses for some datasets. Their licensing guide can help determine the license of a dataset.
        \item For existing datasets that are re-packaged, both the original license and the license of the derived asset (if it has changed) should be provided.
        \item If this information is not available online, the authors are encouraged to reach out to the asset's creators.
    \end{itemize}

\item {\bf New assets}
    \item[] Question: Are new assets introduced in the paper well documented and is the documentation provided alongside the assets?
    \item[] Answer: \answerYes{} 
    \item[] Justification: We provide the documentation along with the code.
    \item[] Guidelines:
    \begin{itemize}
        \item The answer NA means that the paper does not release new assets.
        \item Researchers should communicate the details of the dataset/code/model as part of their submissions via structured templates. This includes details about training, license, limitations, etc. 
        \item The paper should discuss whether and how consent was obtained from people whose asset is used.
        \item At submission time, remember to anonymize your assets (if applicable). You can either create an anonymized URL or include an anonymized zip file.
    \end{itemize}

\item {\bf Crowdsourcing and research with human subjects}
    \item[] Question: For crowdsourcing experiments and research with human subjects, does the paper include the full text of instructions given to participants and screenshots, if applicable, as well as details about compensation (if any)? 
    \item[] Answer: \answerNA{} 
    \item[] Justification: The paper does not involve crowdsourcing nor research with human subjects.
    \item[] Guidelines:
    \begin{itemize}
        \item The answer NA means that the paper does not involve crowdsourcing nor research with human subjects.
        \item Including this information in the supplemental material is fine, but if the main contribution of the paper involves human subjects, then as much detail as possible should be included in the main paper. 
        \item According to the NeurIPS Code of Ethics, workers involved in data collection, curation, or other labor should be paid at least the minimum wage in the country of the data collector. 
    \end{itemize}

\item {\bf Institutional review board (IRB) approvals or equivalent for research with human subjects}
    \item[] Question: Does the paper describe potential risks incurred by study participants, whether such risks were disclosed to the subjects, and whether Institutional Review Board (IRB) approvals (or an equivalent approval/review based on the requirements of your country or institution) were obtained?
    \item[] Answer: \answerNA{} 
    \item[] Justification: The paper does not involve crowdsourcing nor research with human subjects.
    \item[] Guidelines:
    \begin{itemize}
        \item The answer NA means that the paper does not involve crowdsourcing nor research with human subjects.
        \item Depending on the country in which research is conducted, IRB approval (or equivalent) may be required for any human subjects research. If you obtained IRB approval, you should clearly state this in the paper. 
        \item We recognize that the procedures for this may vary significantly between institutions and locations, and we expect authors to adhere to the NeurIPS Code of Ethics and the guidelines for their institution. 
        \item For initial submissions, do not include any information that would break anonymity (if applicable), such as the institution conducting the review.
    \end{itemize}

\item {\bf Declaration of LLM usage}
    \item[] Question: Does the paper describe the usage of LLMs if it is an important, original, or non-standard component of the core methods in this research? Note that if the LLM is used only for writing, editing, or formatting purposes and does not impact the core methodology, scientific rigorousness, or originality of the research, declaration is not required.
    \item[] Answer: \answerNA{} 
    \item[] Justification: The core method development in this research does not involve LLMs as any important, original, or non-standard components.
    \item[] Guidelines:
    \begin{itemize}
        \item The answer NA means that the core method development in this research does not involve LLMs as any important, original, or non-standard components.
        \item Please refer to our LLM policy (\url{https://neurips.cc/Conferences/2025/LLM}) for what should or should not be described.
    \end{itemize}

\end{enumerate}

\end{document}